%% file: bare_jrnl_new_sample4.tex
\documentclass[letterpaper,journal]{IEEEtran}
\usepackage{amsmath,amsfonts}
\usepackage{algorithmic}
\usepackage{algorithm}
\usepackage{array}
\usepackage[caption=false,font=normalsize,labelfont=sf,textfont=sf]{subfig}
\usepackage{textcomp}
\usepackage{stfloats}
\usepackage{url}
\usepackage{verbatim}
\usepackage{graphicx}
\usepackage{cite}
\usepackage{multirow}
\usepackage{xcolor}
\usepackage[table]{xcolor}
\usepackage{makecell}
\usepackage[
    colorlinks=true,
    citecolor=blue,
    linkcolor=red,
]{hyperref}

\begin{document}

\title{Taming Real-World Space-Time Video Super-Resolution with One-Step Diffusion}

\author{Shuoyan~Wei,
        Feng~Li,~\IEEEmembership{Member,~IEEE,}
        Chen~Zhou,
        Runmin~Cong,~\IEEEmembership{Senior Member,~IEEE,}
        Yao~Zhao,~\IEEEmembership{Fellow,~IEEE,}
        and~Huihui~Bai,~\IEEEmembership{Senior Member,~IEEE}
\thanks{Shuoyan Wei, Chen Zhou, Yao Zhao, and Huihui Bai are with the Institute of Information Science, Beijing Jiaotong University, Beijing 100044, China, and Visual Intelligence + X International Cooperation Joint Laboratory of MOE, Beijing 100044, China. (Email: \{shuoyan.wei, chenzhou, yzhao, hhbai\}@bjtu.edu.cn)}
\thanks{Feng Li is with the Innovation School of Artificial Intelligence, Hefei University of Technology, Hefei 230601, China. (Email: fengli@hfut.edu.cn)}
\thanks{Runmin Cong is with the School of Control Science and Engineering, Shandong University, Jinan 250100, China. (Email: rmcong@sdu.edu.cn)}
\thanks{Corresponding author: Feng Li.}
}

\markboth{Journal of \LaTeX\ Class Files,~Vol.~14, No.~8, August~2021}%
{Shell \MakeLowercase{\textit{et al.}}: A Sample Article Using IEEEtran.cls for IEEE Journals}

\maketitle

\begin{abstract}
Diffusion models have demonstrated exceptional success in video super-resolution (VSR), exhibiting powerful capabilities for generating fine-grained details. However, their potential for space-time video super-resolution (STVSR), which necessitates not only recovering realistic high-resolution visual content but also improving the frame rate with coherent temporal dynamics, remains largely underexplored. Moreover, existing STVSR methods predominantly address spatiotemporal upsampling under simple degradation assumptions, thus failing in real-world scenarios with complex unknown degradations. To address these challenges, we propose OSDEnhancer, the first framework that achieves robust STVSR in one-step diffusion. OSDEnhancer begins with a linear initialization to establish essential spatiotemporal structures and adapt the model for one-step reconstruction. It then applies a divide-and-conquer strategy, introducing the temporal coherence (TC) and texture enrichment (TE) LoRAs that progressively specialize in inter-frame dynamics modeling and fine-grained texture recovery, respectively, while collaborating during inference for enhanced overall performance. A bidirectional VAE decoder employs deformable recurrent blocks to leverage the multi-scale structure of the vanilla VAE, enhancing latent-to-pixel reconstruction through joint multi-scale deformable aggregation and inter-frame feature propagation. Experimental results demonstrate that the proposed method attains state-of-the-art performance with superior generalization in real-world scenarios. The code is available at \url{https://github.com/W-Shuoyan/OSDEnhancer}.
\end{abstract}


\section{Introduction}
\IEEEPARstart{V}{ideos} rescaling across spatial and temporal dimensions is widely applied in video streaming and continuous media data~\cite{tian2021self,qiu2023learning,li2024enhanced} to ensure cross-device compatibility, efficient transmission, and storage savings, often at the expense of reduced spatial resolution and temporal frame rates. This necessity has raised the development of video super-resolution (VSR)~\cite{zhang2024realviformer,xu2025videogigagan,shi2025self} and frame interpolation (VFI)~\cite{briedis2025controllable,seo2025bim,hur2025high} techniques, which perform spatial or temporal upscaling as disjoint problems, limiting flexibility in practical applications. 

\begin{figure*}[t]
  \includegraphics[width=\textwidth]{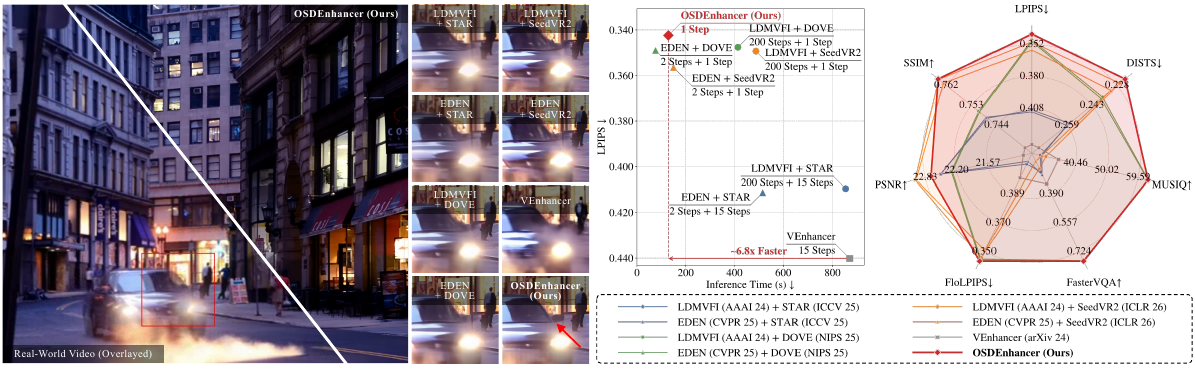}
  \caption{Performance and efficiency comparison on real-world STVSR. OSDEnhancer demonstrates superior reconstruction on interpolated frames over the real-world VideoLQ dataset~\cite{chan2022investigating}, exhibiting clearer structures and details. Moreover, it achieves a  better trade-off between quality and efficiency than state-of-the-art DM-based methods on the real-world MVSR4x dataset~\cite{wang2023benchmark} under generating a 97-frame $1024\times1024$ video with single-frame interpolation, while delivering a $\sim6.8\times$ speedup over the recent DM-based STVSR approach VEnhancer~\cite{he2024venhancer} on an NVIDIA A800 GPU.}
  \label{fig:teaser}
\end{figure*}

Space-time video super-resolution (STVSR)~\cite{xiao2020space, xiang2020zooming, xu2021temporal, zhang2022optical} handles these shortcomings by reconstructing a high-resolution (HR), high-frame-rate (HFR) video from its low-resolution (LR), low-frame-rate (LFR) counterpart within a unified model. Nevertheless, existing methods~\cite{chen2023motif, kim2025bf, zhang2025continuous, wei2025evenhancer} are primarily tailored for ideally known degradations (\emph{e.g.}, bicubic downsampling). This oversimplification renders them fragile when confronted with the complex and heterogeneous degradations in real-world scenarios. Although recent studies~\cite{chan2022investigating,peng2025mitigating, han2025dc, kong2025dam} have increasingly focused on real-world VSR, especially empowered by diffusion models (DMs)~\cite{rombach2022high,esser2024scaling,yang2024cogvideox} for their strong generative capability, the naive cascading of independent VSR and VFI models fails to exploit intrinsic spatiotemporal correlations in video sequences, thus leading to error accumulation and compromised structural fidelity (see Fig.~\ref{fig:teaser}). Therefore, effectively harnessing DMs to conquer real-world degradations while preserving realistic restoration in a unified STVSR framework remains a pivotal challenge. 

On the other hand, the protracted iterative sampling process inherent in DMs~\cite{zhou2024upscale, wang2025seedvr,jain2024video, zhu2025generative} incurs prohibitive computational overhead, especially for long sequences and resource-constrained deployments. VEnhancer~\cite{he2024venhancer} pioneers the generative space-time enhancement method in a video diffusion model (VDM) and reduces the sampling trajectory to 15 steps. However, as shown in Fig.~\ref{fig:teaser}, it still suffers from severe latency. While recent VSR methods~\cite{chen2025dove, liu2025ultravsr, sun2025one} have accelerated DMs to extreme one-step inference, extending this strategy to STVSR encounters a fundamental barrier, where the simultaneous absence of intermediate frames and HR spatial details induces compounded ambiguity in both space and time. 

In this work, we propose OSDEnhancer, a novel DM-based framework that transcends prior limitations for real-world STVSR in a one-step sampling paradigm. Built upon pretrained VDMs~\cite{hong2022cogvideo,yang2024cogvideox}, OSDEnhancer begins with a linear initialization to establish essential spatiotemporal structures aligned with the target video sequences and adapts the model to enable one-step reconstruction with arbitrary spatiotemporal upscaling. Then, we devise a divide-and-conquer scheme that disentangles STVSR into complementary temporal coherence (TC) and texture enrichment (TE) adaptations, equipped with corresponding specialized low-rank adapters (LoRAs)~\cite{hu2021lora} that share the same pretrained diffusion transformer (DiT) but undergo progressive fine-tuning dedicated to strong generative capability. To rigorously reinforce inter-frame coherence, we leverage temporal residuals to guide the TC-LoRA toward regions of pronounced inter-frame variations, enhancing its capacity to model temporal dynamics effectively. Subsequently, recognizing that the inherent compression characteristic of the variational autoencoder (VAE) in DMs can suppress the recovery of high-frequency details, we implement TE-LoRA operating in pixel space to improve fine-grained textures. Finally, we introduce a bidirectional deformable VAE decoder that leverages the inherent multi-scale structure of the vanilla VAE, performing recurrent deformable inter-frame propagation within each scale and alternating across adjacent scales for efficient bidirectional alignment, while propagating low-scale offsets to higher scales to facilitate precise motion compensation and globally coherent latent-to-pixel reconstruction. As illustrated in Fig.~\ref{fig:teaser}, extensive experiments demonstrate that the proposed OSDEnhancer significantly outperforms existing state-of-the-art methods under real-world degradations while maintaining excellent efficiency among DM approaches. The main contributions are as follows:
\begin{itemize}
\item We propose OSDEnhancer, to the best of our knowledge, the first DM-based STVSR approach to achieve one-step inference. Extensive experiments validate its superiority under complex degradations.
\item We propose a divide-and-conquer adaptation scheme with dedicated TC- and TE-LoRAs that are progressively fine-tuned on a shared DiT backbone to improve temporal coherence and texture richness.
\item A bidirectional deformable VAE decoder is designed with recurrent inter-frame deformable compensation across adjacent scales to strengthen spatiotemporal dependency in latent-to-pixel reconstruction.
\end{itemize}

\section{Related Work}

\subsection{Space-Time Video Super-Resolution}
Space-time video super-resolution (STVSR)~\cite{shechtman2002increasing} unifies the objectives of VSR~\cite{li2018video,wen2022video,chan2022investigating,qiu2023learning} and VFI~\cite{bao2019depth,shen2020video,jain2024video,seo2025bim}, aiming to increase both spatial and temporal resolutions simultaneously from LR and LFR videos. Early methods~\cite{kim2020fisr,xiang2020zooming,hu2023store} mainly focus on fixed discrete scales in space and time. FISR~\cite{kim2020fisr} introduces the first joint VFI and VSR framework with multi-scale temporal regularization. STARNet~\cite{haris2020space} and SAFA~\cite{huang2024scale} apply traditional optical flow to perform temporal feature compensation and aggregation. Motivated by the effectiveness of deformable convolution~\cite{zhu2019deformable} in VSR~\cite{tian2020temporally,wang2019edvr}, some methods~\cite{xiang2020zooming,xu2021temporal,hu2022spatial,hu2023cycmunet+} leverage deformable sampling to interpolate missing intermediate frame features. RSTT~\cite{geng2022rstt} incorporates the temporal interpolation and spatial super-resolution modules for STVSR without explicit motion compensation. Later methods extend beyond fixed upscaling factors, enabling arbitrary-scale STVSR through continuous video implicit neural representations (INR)~\cite{chen2022videoinr,chen2023motif,wei2025evenhancer,zhang2025space}. VideoINR~\cite{chen2022videoinr} is the pioneering method that learns respective continuous spatial and temporal INRs. EvEnhancer~\cite{wei2025evenhancer} learns a unified video INR with auxiliary event streams. BF-STVSR~\cite{kim2025bf} employs a B-spline mapper for temporal motion representations and a Fourier mapper to capture fine-grained spatial details. V$^3$~\cite{becker2025continuous} models continuous video representations in a 3D Fourier field, effectively reducing runtime and memory footprint. Most recently, VEnhancer~\cite{he2024venhancer} adopts pretrained VDM~\cite{zhang2023i2vgen} as the video generative prior and builds a unified DM-based framework to support flexible spatial and temporal scales. However, current approaches are tailored exclusively to synthetic bicubic degradation, leaving them ill-suited for practical applications. Instead, this work generalizes STVSR to complex real-world degradations and achieves plausible reconstruction at flexible scales.

\subsection{One-Step Diffusion Model Acceleration}
While DMs have demonstrated impressive capabilities across various generative tasks, their reliance on iterative denoising steps imposes prohibitive computational costs and inference latency. To address the inefficiency problem, considerable efforts have been devoted to reducing the inference steps for DM acceleration, with recent works pushing toward the extreme of one-step diffusion, such as diffusion distillation~\cite{yin2024one,wang2024sinsr,liu2023instaflow} and adversarial post-training~\cite{lin2025diffusion,sun2025pixel}. For instance, UltraVSR~\cite{liu2025ultravsr} introduces degradation-aware reconstruction scheduling to achieve one-step reconstruction through spatiotemporal joint distillation. DLoRAL~\cite{sun2025one} draws inspiration from one-step single-image super-resolution~\cite{sun2025pixel} and presents a dual LoRA learning framework for VSR. SeedVR2~\cite{wang2025seedvr2} performs adversarial training with a pretrained DiT as initialization to tackle the one-step video restoration problem. FlashVSR~\cite{zhuang2025flashvsr} constructs a three-stage distillation pipeline to pursue real-time streaming VSR. DOVE~\cite{chen2025dove} devises a latent-pixel training strategy that adapts the pretrained VDM to one-step VSR. RDVFI~\cite{marealtime} solves large complex motions using high-order continuous pixel trajectories, thus enabling one-step sampling based on latent VDM for VFI. Despite extensive advances in individual VSR and VFI tasks, their extension to joint STVSR remains entirely unexplored. In this work, the proposed OSDEnhancer pioneers a distillation-free STVSR framework that progressively enhances temporal coherence and texture details with one-step inference.

\begin{figure*}[t]
  \includegraphics[width=\textwidth]{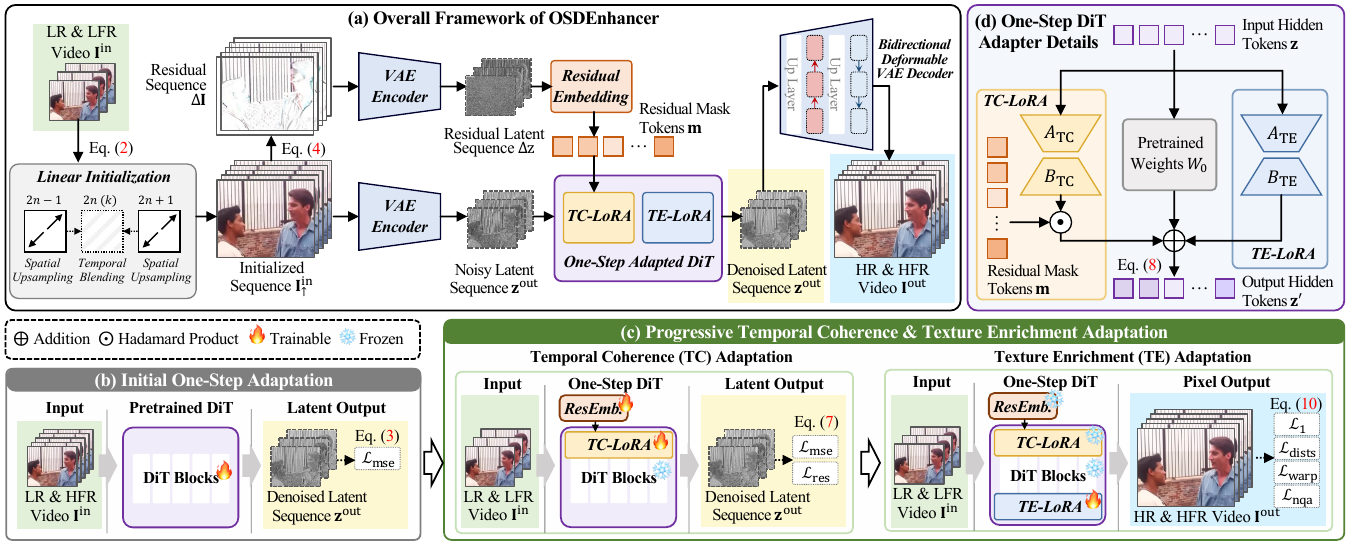}
 \caption{Overall architecture and adaptation pipeline of OSDEnhancer. (a) The overall framework generates an HR and HFR video from an LR and LFR input via one-step diffusion. (b) Initial one-step adaptation enables the pretrained multi-step DiT to operate under the one-step modeling paradigm. (c) The divide-and-conquer adaptation scheme progressively learns the TC- and TE-LoRAs, enabling collaborative modeling of temporal dynamics and texture enrichment. (d) The finally one-step adapted DiT consists of complementary TC- and TE-LoRAs.}
  \label{fig:framework}
\end{figure*}

\section{Method}

\subsection{Preliminary}
\label{sec:3-1}
OSDEnhancer is built upon the pretrained video diffusion model CogVideoX~\cite{yang2024cogvideox}, which utilizes a 3D causal VAE to compress a video into latent code $\mathbf{z}$ and a DiT to perform the diffusion process. In the forward process, a clean latent code $\mathbf{z}_0$ is progressively noised into $\mathbf{z}_t$ by Gaussian noise $\epsilon$: $\mathbf{z}_t=\sqrt{\bar{\alpha}_{t}}\mathbf{z}_0+\sqrt{1-\bar{\alpha}_{t}}\epsilon$, where $\bar{\alpha}_{t}$ is the predefined schedule factor at timestep $t$. Given that the input video $\mathbf{I}^\mathrm{in}$ already provides structural information rather than pure noise, following prior VSR methods~\cite{liu2025ultravsr, sun2025one, chen2025dove}, we treat the latent sequence $\mathbf{z}^{\mathrm{in}}$ derived from $\mathbf{I}^\mathrm{in}$ as the starting point of the denoising process and produce the target video sequence $\mathbf{z}^{\mathrm{out}}$. Under the $v$-prediction formulation of CogVideoX, the denoising process is expressed as
\begin{equation}
\label{eqn:01}
\mathbf{z}^{\mathrm{out}}=\sqrt{\bar{\alpha}_{t}}\mathbf{z}^{\mathrm{in}}-\sqrt{1-\bar{\alpha}_{t}}\textbf{v}_{\theta}(\mathbf{z}^{\mathrm{in}}, \mathbf{c}, t), 
\end{equation}
where $\textbf{v}_{\theta}$ is the predicted velocity under the condition $\mathbf{c}$.

\subsection{Overall framework}
\label{sec:3-2}
The overall framework of OSDEnhancer is illustrated in Fig.~\ref{fig:framework}(a), which enables one-step STVSR for real-world degradations. Given an LR and LFR video sequence $\mathbf{I}^{\mathrm{in}}=\{I_{2n-1}^{\mathrm{in}}\}_{n=1}^{N}$, our goal is to generate an HR and HFR version $\textbf{I}^{\mathrm{out}}=\{I_{1}^{\mathrm{out}}, I_{2(1)}^{\mathrm{out}},..., I_{2(K)}^{\mathrm{out}}, I_{3}^{\mathrm{out}},...,I_{2N-1}^{\mathrm{out}}\}$,
where $K$ denotes the number of temporally interpolated frames between two consecutive input frames. For example, $\{I_{2n(k)}^{\mathrm{in}}\}_{k=1}^{K}$ indicates the interpolated frames between $I_{2n-1}^{\mathrm{in}}$ and $I_{2n+1}^{\mathrm{in}}$. 

Unlike VSR, directly implementing STVSR on a pretrained DM presents a distinct challenge. The input in STVSR is simultaneously devoid of high-frequency spatial details and temporally absent intermediate frames. According to Eq.~\eqref{eqn:01}, the denoising process from $\mathbf{z}^{\mathrm{in}}$ to $\mathbf{z}^{\mathrm{out}}$ inherently preserves the dimensionality of the latent space. To obtain output frames at the target resolution, VSR methods~\cite{liu2025ultravsr, sun2025one, chen2025dove} typically derive $\mathbf{z}^{\mathrm{in}}$ by spatially upsampling the input frames before VAE encoding. However, STVSR demands not only spatial scale alignment but also temporal length completion. To bridge this gap, in OSDEnhancer, we first conduct a simple linear initialization on $\mathbf{I}^\mathrm{in}$ before VAE encoding, generating an aligned sequence $\mathbf{I}_{\uparrow}^{\mathrm{in}}$ used to adapt the original multi-step DiT to be one-step. Then, we calculate the temporal residuals from $\mathbf{I}_{\uparrow}^{\mathrm{in}}$, resulting in ${\Delta\mathbf{I}}$, which are together fed into the VAE encoder to produce corresponding $\mathbf{z}^{\mathrm{in}}$ and ${\Delta}\mathbf{z}$. We design the temporal coherence (TC) and texture enrichment (TE) LoRAs that are progressively fine-tuned on a shared DiT backbone to improve temporal coherence and textures, yielding the latent sequence $\mathbf{z}^{\mathrm{out}}$ decoded by our bidirectional deformable VAE decoder to produce the final result $\mathbf{I}^{\mathrm{out}}$.

\subsection{Initial One-Step Adaptation}
\label{sec:3-3}
In STVSR, the simultaneous spatial and temporal degradations render it far harder to fine-tune VDMs for high-fidelity restoration than existing video generation tasks~\cite{ho2022video,zhang2023i2vgen,yang2024cogvideox}. To ease the difficulty, we first present a simple linear initialization strategy to establish essential spatiotemporal
structures aligned with the target video sequences. It replenishes the temporal dimension by performing weighted blending for the elements $\{I_{\uparrow,2n(k)}^{\mathrm{in}}\}_{k=1}^{K}$ at potential intermediate frame positions in the aligned sequence $\mathbf{I}_{\uparrow}^{\mathrm{in}}$ according to their relative temporal distances to the adjacent keyframes $I_{2n-1}^{\mathrm{in}}$ and $I_{2n+1}^{\mathrm{in}}$. This can be formulated as 
\begin{equation}
\label{eqn:03}
\left\{
\begin{aligned}
I_{\uparrow,2n-1}^{\mathrm{in}}&=\texttt{Up}(I_{2n-1}),\\
I_{\uparrow,2n(k)}^{\mathrm{in}}&=\texttt{Up}(\frac{K+1-k}{K+1}I_{2n-1}^{\mathrm{in}}+\frac{k}{K+1}I_{2n+1}^{\mathrm{in}}),
\end{aligned}
\right.
\end{equation}
where $\texttt{Up}(\cdot)$ denotes the spatial upsampling operation. The weighted blending supplies temporal cues for flexible multi-frame inference within a single forward pass. By varying $\texttt{Up}(\cdot)$ and the number of interpolated frames $K$, this initialization naturally supports arbitrary spatiotemporal upscaling.

Then, we adapt the pretrained multi-step DiT to enable it to model spatiotemporal correlations effectively in a single step (Fig.~\ref{fig:framework}(b)). In this stage, the adaptation is intentionally limited to spatial degradation alone, isolating it from temporal complexity to ensure stable and tractable convergence. Accordingly, we fine-tune the DiT model on paired data $\{\mathbf{I}^{\mathrm{in}}, \mathbf{I}^{\mathrm{gt}}\}$ from the high-quality (HQ) video dataset, where $\mathbf{I}^{\mathrm{in}}$ contains only spatial degradation and $\mathbf{I}^{\mathrm{gt}}$ is the corresponding ground-truth (GT) sequence. As the latent space compactly encodes spatiotemporal information, we employ the MSE loss $\mathcal{L}_{\mathrm{mse}}$ as the regression objective in the latent space:
\begin{equation}
\label{eqn:05}
\mathcal{L}_{\mathrm{initial}}=\mathcal{L}_{\mathrm{mse}}(\mathbf{z}^{\mathrm{out}},\mathbf{z}^{\mathrm{gt}}),
\end{equation}
where $\mathbf{z}^{\mathrm{gt}}$ is the latent code of $\mathbf{I}^{\mathrm{gt}}$.

\subsection{Progressive Temporal Coherence and Texture Enrichment Adaptation}
\label{sec:3-4}
The initial adaptation under spatial degradation may fail in inter-frame motion synthesis and texture recovery, particularly in large motions or occlusions. To solve this issue, we propose a divide-and-conquer scheme that progressively guides the model from latent-space temporal dynamics modeling to pixel-space texture enrichment through complementary temporal coherence (TC) and texture enrichment (TE) LoRAs, as illustrated in Fig.~\ref{fig:framework}(c).

\textbf{TC Adaptation}. We introduce the TC-LoRA, which specializes in modeling spatiotemporal structures under temporal degradation. To obtain this LoRA, we train it with input video sequences $\mathbf{I}^{\mathrm{in}}$ derived from the HFR video dataset through spatial and temporal degradation. Besides, we compute the temporal residuals ${\Delta\mathbf{I}}$ between spatially upsampled keyframes $I_{\uparrow,2n-1}^{\mathrm{in}}$ and $I_{\uparrow,2n+1}^{\mathrm{in}}$ (Eq.~(\ref{eqn:03})) as an auxiliary signal to further reinforce inter-frame coherence, defined as 
\begin{equation}
\label{eqn:04}
\left\{
\begin{aligned}
{\Delta}I_{2n-1}&=\mathbf{0},\\
{\Delta}I_{2n(k)}&=I_{\uparrow,2n+1}^{\mathrm{in}}-I_{\uparrow,2n-1}^{\mathrm{in}}.
\end{aligned}
\right.
\end{equation}

Then, the adapter learns low-rank decomposition matrices $A_{\mathrm{TC}}$ and $B_{\mathrm{TC}}$, with their output adaptively modulated through a Hadamard product by residual mask tokens $\mathbf{m}$ obtained from the residual latent sequence ${\Delta}\mathbf{z}$ via 3D patch embedding. By explicitly incorporating these motion variation cues, the TC-LoRA effectively captures inter-frame variations, facilitating the synthesis of non-linear motion and temporally coherent content between adjacent keyframes. This process is formulated as
\begin{equation}
\label{eqn:06}
\mathbf{z}'=W_{0}\mathbf{z}+\mathbf{m} \odot B_{\mathrm{TC}} A_{\mathrm{TC}} \mathbf{z},
\end{equation}
where $\odot$ denotes the Hadamard product. $\mathbf{z}$ and $\mathbf{z}'$ represent the input hidden feature tokens and the modulated output of the network component, respectively. $W_0$ denotes the pretrained weight. The residual mask $\mathbf{m}$ acts as a spatiotemporal gate that controls attention on motion-intensive areas while preserving the static structural integrity of the backbone. 

During this stage, all network parameters remain frozen except for $A_{\mathrm{TC}}$, $B_{\mathrm{TC}}$, and the residual embedding modules for $\Delta\mathbf{z}$. In addition to the standard MSE loss, we introduce a residual loss $\mathcal{L}_{\mathrm{res}}$ to ensure temporal coherence across frames. This term constrains the distance between consecutive residuals in the predicted latent sequence $\mathbf{z}^{\mathrm{out}}$ and those in the GT latent sequence $\mathbf{z}^{\mathrm{gt}}$, formulated as
\begin{equation}
\label{eqn:07a}
\mathcal{L}_\mathrm{res}(\mathbf{z}^{\mathrm{out}},\mathbf{z}^{\mathrm{gt}})=\frac{1}{J_\mathbf{z}-1}\sum_{j=1}^{J_\mathbf{z}-1}||(z_{j+1}^\mathrm{out}-z_{j}^\mathrm{out})-(z_{j+1}^\mathrm{gt}-z_{j}^\mathrm{gt})||,
\end{equation}
where $z_{j}^\mathrm{out}$ and $z_{j}^\mathrm{gt}$ denote the $j$-th latent frame in their respective sequences $\mathbf{z}^{\mathrm{out}}$ and $\mathbf{z}^{\mathrm{gt}}$, and $J_\mathbf{z}$ represents the latent sequence length. The total objective for the TC adaptation is defined as
\begin{equation}
\label{eqn:07b}
\mathcal{L}_{\mathrm{TC}}=\mathcal{L}_{\mathrm{mse}}(\mathbf{z}^{\mathrm{out}},\mathbf{z}^{\mathrm{gt}}) + \lambda_{\mathrm{res}}\mathcal{L}_{\mathrm{res}}(\mathbf{z}^{\mathrm{out}},\mathbf{z}^{\mathrm{gt}}),
\end{equation}
where $\lambda_{\mathrm{res}}$ is a weight that balances the residual loss. 

\textbf{TE Adaptation}. While the latent space learning in the TC adaptation facilitates efficient inter-frame dynamics modeling, the inherent spatiotemporal compression of the latent sequence in the DM can hinder fine-grained texture restoration. Directly fine-tuning all DiT parameters on HQ sequences is essential to recover these details, but high-dimensional data involves prohibitive computational and memory overhead. To circumvent this, we introduce the TE-LoRA, transitioning the learning paradigm from latent space to pixel space. By augmenting the DiT with additional low-rank matrices $A_{\mathrm{TE}}$ and $B_{\mathrm{TE}}$, this component is specifically optimized with pixel-level supervision to refine textures upon the previously established structures. Following previous VDM works~\cite{yang2024cogvideox,chen2025dove,wan2025wan}, we jointly employ HQ image and video datasets at this stage to ensure rich textures and fine details. As shown in Fig.~\ref{fig:framework}(d), together with the TC-LoRA, we can yield a joint output formulated as
\begin{equation}
\label{eqn:08}
\mathbf{z}'=W_{0}\mathbf{z}+\mathbf{m} \odot B_{\mathrm{TC}} A_{\mathrm{TC}} \mathbf{z} + B_{\mathrm{TE}} A_{\mathrm{TE}} \mathbf{z}.
\end{equation}
In this manner, the TC- and TE-LoRAs synergistically model inter-frame dynamics and fine-grained details. 

The optimization at this stage incorporates losses from several aspects. Specifically, we employ the L1 loss $\mathcal{L}_{1}$ and DISTS loss $\mathcal{L}_{\mathrm{dists}}$~\cite{ding2020image} for structural and textural fidelity. To further enforce the temporal consistency of enhanced textures, a self-supervised optical-flow warping loss $\mathcal{L}_{\mathrm{warp}}$~\cite{lai2018learning} is applied to the high-frequency component $H_{j}^\mathrm{out}$ of
the output frame $I_{j}^\mathrm{out}$ in the sequence $\mathbf{I}^\text{out}$, defined as
\begin{equation}
\label{eqn:09a}
\mathcal{L}_\mathrm{warp}(\mathbf{I}^{\mathrm{out}})=\frac{1}{J_\mathbf{I}-1}\sum_{j=1}^{J_\mathbf{I}-1}\frac{||\hat{M}_j\odot(\hat{O}_{j}(H_{j}^\mathrm{out})-H_{j+1}^\mathrm{out})||}{||\hat{M}_j||+\xi},
\end{equation}
where $\hat{O}_{j}$ denotes the warping operator induced by the optical flow from $I_{j}^\mathrm{out}$ to $I_{j+1}^\mathrm{out}$, while $\hat{M}_j$ is the non-occlusion mask constructed via forward–backward consistency~\cite{wang2018occlusion} to ignore occluded and out-of-bounds regions. $\xi$ is a small constant introduced to avoid division by zero, and $J_\mathbf{I}$ is the output sequence length. Additionally, a no-reference quality assessment loss $\mathcal{L}_{\mathrm{nqa}}$~\cite{ke2021musiq, zhang2025augmenting} is incorporated as a regularization term to further enhance the perceptual quality of the synthesized content. The overall loss for the TE adaptation is formulated as
\begin{equation}
\label{eqn:09}
\begin{aligned}
\mathcal{L}_{\mathrm{TE}}
= {} & \mathcal{L}_{1}(\mathbf{I}^{\mathrm{out}}, \mathbf{I}^{\mathrm{gt}})
+ \lambda_{\mathrm{dists}} \mathcal{L}_{\mathrm{dists}}(\mathbf{I}^{\mathrm{out}}, \mathbf{I}^{\mathrm{gt}}) \\
& + \lambda_{\mathrm{warp}} \mathcal{L}_{\mathrm{warp}}(\mathbf{I}^{\mathrm{out}})
- \lambda_{\mathrm{nqa}} \mathcal{L}_{\mathrm{nqa}}(\mathbf{I}^{\mathrm{out}}),
\end{aligned}
\end{equation}
where $\lambda_{\mathrm{dists}}$, $\lambda_{\mathrm{warp}}$, and $\lambda_{\mathrm{nqa}}$ are loss weights.

\begin{figure}[t]
  \includegraphics[width=\columnwidth]{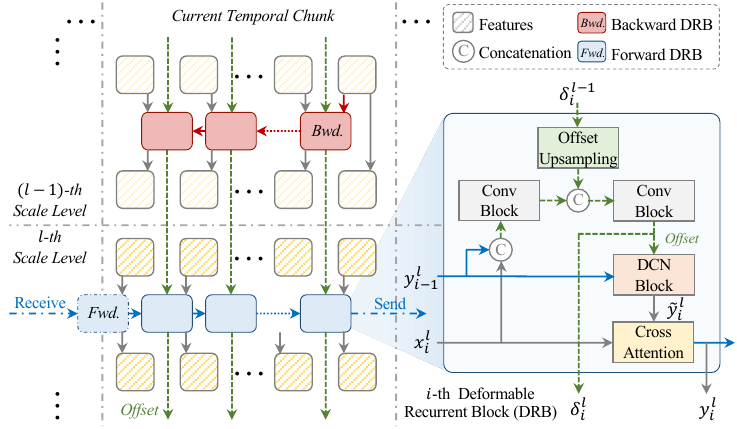}
  \caption{Illustration of the bidirectional deformable VAE decoder. Deformable recurrent blocks (DRBs) integrated into the upsampling layers of a 3D causal VAE decoder, enabling multi-scale cross-frame compensation.}
  \label{fig:vae}
\end{figure}

\subsection{Bidirectional Deformable VAE Decoder}
\label{sec:3-5}
Most VDMs~\cite{ho2022video,yang2024cogvideox,wan2025wan} apply 3D VAE with temporally causal convolutions to capture inter-frame dependencies. In STVSR, the strict temporal causality can limit effective global compensation from interpolated intermediate frames during decoding. To overcome this constraint, we propose a bidirectional deformable VAE decoder, as shown in Fig.~\ref{fig:vae}, which integrates deformable recurrent blocks (DRBs) into the upsampling layers of the vanilla 3D causal VAE decoder to learn multi-scale deformable aggregation and inter-frame feature propagation. 

At each $l$-th scale level, for the $i$-th DRB, the output $y_{i-1}^{l}$ from the previous $(i-1)$-th DRB is first paired with the current input $x_i^{l}$ to estimate the deformation offset through a convolutional block. Inspired by the PCD network~\cite{wang2019edvr}, we further propagate the upsampled offset $\delta_{i}^{l-1}$ at the lower scale level $(l-1)$ to the current level to obtain a more reliable and expressive offset $\delta_{i}^{l}$. Then, $\delta_{i}^{l}$ is used in a deformable convolution block $\texttt{DCN}(\cdot)$~\cite{zhu2019deformable} applied to $y_{i-1}^{l}$, producing the aligned feature $\tilde{y}_{i-1}^{l}$ with respect to $x_{i}^{l}$. Finally, a cross-attention $\texttt{attn}(\cdot)$ is employed to compensate the current $x_{i}^{l}$ with information propagated from the neighboring aligned feature $\tilde{y}_{i-1}^{l}$ to get $y_{i}^{l}$.
 This process can be formulated as
\begin{equation}
\label{eqn:10}
\begin{aligned}
&\delta_{i}^{l} = \texttt{conv}(\texttt{conv}(x_{i}^{l},y_{i-1}^{l}), \texttt{Up}(\delta_{i}^{l-1})), \\
&\tilde{y}_{i-1}^{l} = \texttt{DCN}(y_{i-1}^{l},\delta_{i}^{l}), \quad y_{i}^{l} = \texttt{attn}(x_{i}^{l},\tilde{y}_{i-1}^{l}),
\end{aligned}
\end{equation}
where $\texttt{conv}(\cdot,\cdot)$ denotes the convolution operation. The recurrent propagation within each scale level is performed in a single temporal direction, while the propagation direction is reversed at the next scale level, enabling the preservation of the original temporal chunking for long video processing. Forward information is propagated globally, whereas backward information is restricted to temporal chunks. In this way, we can achieve efficient bidirectional compensation with reduced latency and controlled error accumulation.

During training, we optimize only the DRBs on the same dataset as in the initial adaptation using the same reconstruction loss and perception loss adopted in the vanilla VAE~\cite{yang2024cogvideox}, while the residual loss in Eq.~\eqref{eqn:07a} is further applied in the pixel space to enhance inter-frame consistency.

\input{Tables/Table1v2}
\input{Tables/Table2v2}

\section{Experiments}
\subsection{Experimental Settings}
\textbf{Datasets.}
To support our progressive adaptation scheme, we employ HQ-VSR~\cite{chen2025dove}, Adobe240~\cite{su2017deep}, and DIV2K~\cite{cai2019toward} (with images duplicated to match video length) as the HQ video, HFR video, and HQ image datasets, respectively. Spatial degradations are synthesized using the RealBasicVSR~\cite{chan2022investigating} pipeline, and temporal degradations are conducted by interval sampling, while the original data serve as GT. In the TC adaptation, temporal degradation is introduced by uniformly sampling with frame intervals of 1, 2, 4, 8, or 16. In the TE adaptation, since the HQ video dataset has limited frame rates, we apply mild temporal degradation by uniformly sampling with frame intervals of 1 or 2. For evaluation, we comprehensively consider both synthetic and real-world datasets. The synthetic benchmarks include UDM10~\cite{tao2017detail}, SPMCS~\cite{yi2019progressive}, YouHQ40~\cite{zhou2024upscale}, and GoPro~\cite{nah2017deep}, while the real-world datasets comprise MVSR4x~\cite{wang2023benchmark} and VideoLQ~\cite{chan2022investigating}, where synthetic degradation processes are consistent with the training settings. Following previous STVSR work~\cite{kim2025bf}, we evaluate STVSR performance at multiple spatiotemporal scales on the GoPro dataset with high frame rates. For other datasets with lower frame rates, we only consider single-frame interpolation with $4\times$ spatial upscaling, unless otherwise specified.

\textbf{Implementation Details.}
We adopt CogVideoX1.5-5B~\cite{yang2024cogvideox} as the pretrained backbone of OSDEnhancer. Following~\cite{chen2025dove}, empty text embeddings are consistently used as conditions to stabilize generation and avoid redundant text encoding overhead, while the diffusion timestep is set to $t=399$. Training is conducted on 4 NVIDIA RTX PRO 6000 GPUs with a batch size of 4. We first pretrain the bidirectional deformable VAE decoder for stable latent-to-pixel reconstruction, using 33-frame video sequences at $256\times256$ for 5,000 iterations with a learning rate of $1\times10^{-4}$. In the initial and TC adaptations, we train on 33-frame video sequences at $320\times640$, with 15,000 iterations at a learning rate of $2\times10^{-5}$ in the initial adaptation and 10,000 iterations at a learning rate of $1\times10^{-4}$ with $\lambda_{\mathrm{res}}=1$ in the TC adaptation. In the TE adaptation, we use 9-frame video/image sequences at $320\times320$ and train for 5,000 iterations with a learning rate of $5\times10^{-5}$. The loss weights $\lambda_{\mathrm{dists}}$, $\lambda_{\mathrm{warp}}$, and $\lambda_{\mathrm{nqa}}$ are set to 1, 0.05, and 0.05, respectively. All stages are optimized using AdamW~\cite{loshchilov2017fixing} with $\beta_1=0.9$ and $\beta_2=0.95$. The TC- and TE-LoRAs are injected into the query, key, value, and output projections of 3D attention, as well as the projection layers of the feed-forward networks in the DiT, while the TE-LoRA is further applied to the final output projection layer. Both LoRAs use rank $r=128$ with a scaling factor $\alpha=128$.

\textbf{Evaluation Metrics.}
We comprehensively evaluate STVSR performance using a diverse set of quality metrics, including PSNR and SSIM~\cite{wang2004image} for pixel-level fidelity, LPIPS~\cite{zhang2018unreasonable} and DISTS~\cite{ding2020image} for perceptual quality, and FloLPIPS~\cite{danier2022flolpips} for perceptual similarity with respect to temporal consistency. We also employ no-reference image quality assessment (IQA) metrics, including MUSIQ~\cite{ke2021musiq} and CLIP-IQA~\cite{wang2023exploring}, together with no-reference video quality assessment (VQA) metrics, namely FasterVQA~\cite{wu2023neighbourhood} and DOVER~\cite{wu2023exploring}. Since the ground truth of the real-world dataset VideoLQ~\cite{chan2022investigating} is unavailable, we only report no-reference IQA and VQA metrics. 

\subsection{Comparison with State-of-the-Art Methods}
We compare our OSDEnhancer with state-of-the-art STVSR methods: 1) two-stage cascading methods that integrate VFI methods including LDMVFI~\cite{danier2024ldmvfi} and EDEN~\cite{zhang2025eden}, with VSR methods including STAR~\cite{xie2025star}, DOVE~\cite{chen2025dove}, and SeedVR2-7B~\cite{wang2025seedvr2}; and 2) one-stage unified STVSR methods including VideoINR~\cite{chen2022videoinr}, MoTIF~\cite{chen2023motif}, BF-STVSR~\cite{kim2025bf}, STNO~\cite{zhang2025space}, V$^3$~\cite{becker2025continuous}, and DM-based VEnhancer~\cite{he2024venhancer}. Since VideoINR, MoTIF, BF-STVSR, STNO, and V$^3$ are non-DM methods originally developed under the bicubic downsampling assumption, we retrain them on the same datasets and degradation settings as ours to ensure a fair comparison.

\textbf{Quantitative Comparison.}
The quantitative results are reported in Table~\ref{tab:1}. Generally, OSDEnhancer shows the best performance in both IQA and VQA metrics across most datasets, demonstrating strong superiority. While SeedVR2~\cite{wang2025seedvr2} outperforms in some perceptual metrics on YouHQ40~\cite{zhou2024upscale}, its performance is highly inconsistent across other datasets due to the instability caused by distillation. Moreover, we can see that OSDEnhancer performs comparably against existing methods on the real-world MVSR4x~\cite{wang2023benchmark} and VideoLR~\cite{chan2022investigating} datasets, further validating its generalization ability and robustness.

\begin{figure*}[t]
  \includegraphics[width=\textwidth]{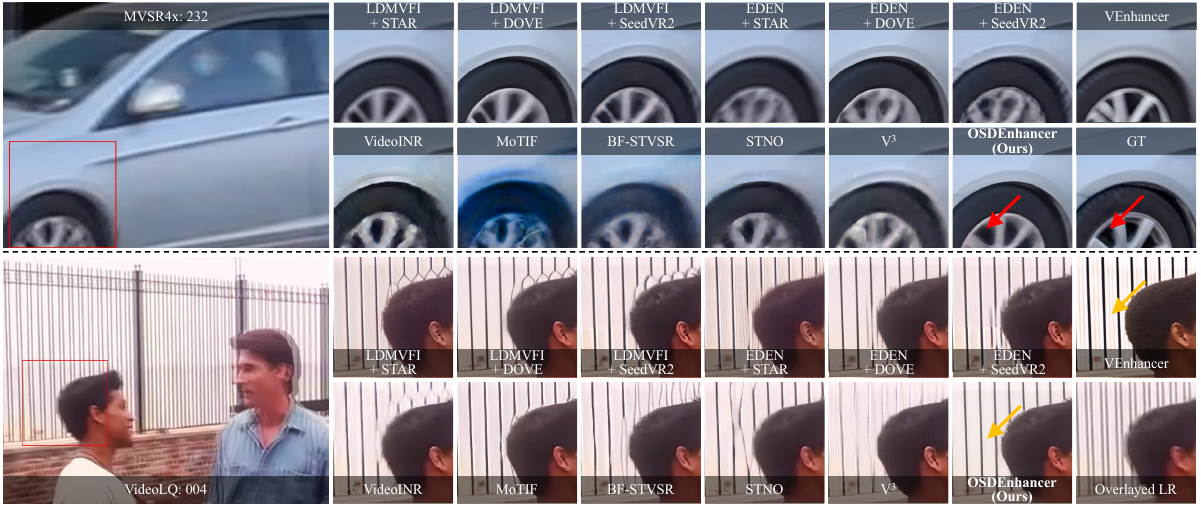}
  \caption{Qualitative comparison on real-world degraded videos from MVSR4x~\cite{wang2023benchmark} and VideoLQ~\cite{chan2022investigating}. Left: overlay of adjacent LR frames.}
  \label{fig:visual1}
\end{figure*}

\begin{figure*}[t]
  \includegraphics[width=\textwidth]{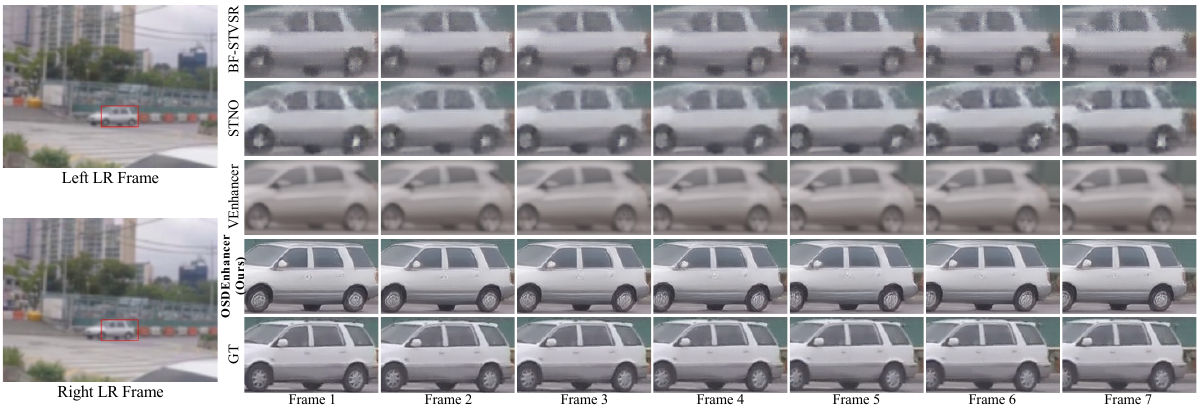}
  \caption{Qualitative comparison of STVSR on GoPro~\cite{nah2017deep} with $8\times$ spatial upscaling and 5-frame interpolation (frames 2–6 are interpolated).}
  \label{fig:visual2}
\end{figure*}

\begin{figure*}[t]
  \includegraphics[width=\textwidth]{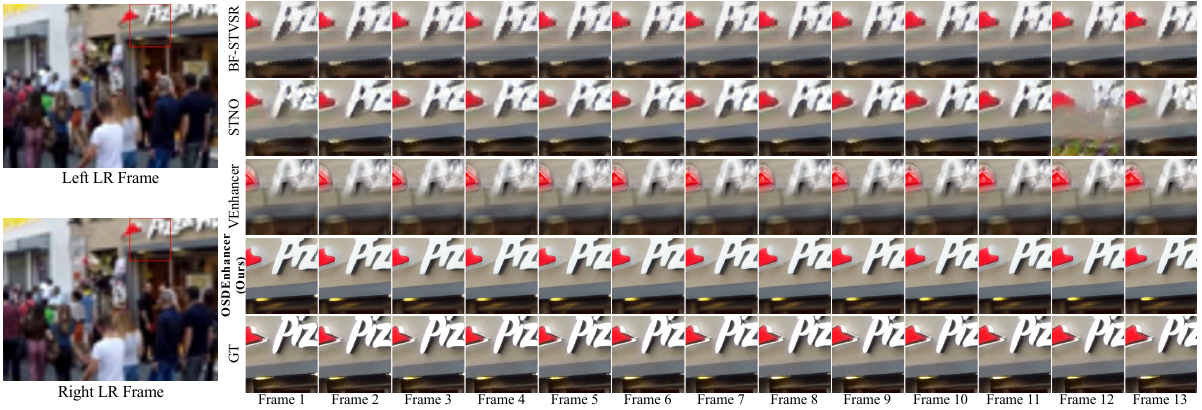}
  \caption{Qualitative comparison of STVSR on GoPro~\cite{nah2017deep} with $12\times$ spatial upscaling and 11-frame interpolation (frames 2–12 are interpolated).}
  \label{fig:visual3}
\end{figure*}

Notably, in contrast to most DM-based methods that are restricted to fixed upscaling factors, our OSDEnhancer allows for arbitrary spatiotemporal upscaling, where the results across different spatiotemporal scales on the GoPro dataset~\cite{nah2017deep} with advanced STVSR methods are reported in Table~\ref{tab:2}. It can be observed that OSDEnhancer not only achieves the best performance for multi-frame reconstruction within the training distribution (temporal scale is $8\times$, spatial scale is $4\times$), but also maintains superior perceptual quality and temporal consistency on out-of-distribution spatiotemporal scales, as evidenced by the best LPIPS and FloLPIPS.

\begin{figure}[t]
  \centering
  \includegraphics[width=\columnwidth]{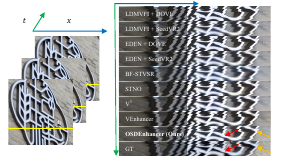}
  \caption{Temporal profiles on the real-world MVSR4x dataset~\cite{wang2023benchmark}. We select a row (yellow lines) and observe the changes across time.}
  \label{fig:temporal}
\end{figure}

\textbf{Qualitative Comparison.}
Fig.~\ref{fig:visual1} presents visual comparisons of real-world datasets under single-frame interpolation with $4\times$ spatial upscaling. Faced with non-linear situations such as wheel rotation, OSDEnhancer reconstructs sharper structures and more faithful details, whereas other methods suffer from heavy texture distortions and blurring. Even VEnhancer~\cite{he2024venhancer}, which performs STVSR in 15 sampling steps, still produces reconstructed frames with unsatisfactory artifacts appearing in the region indicated by the orange arrow. Moreover, Fig.~\ref{fig:visual2} and Fig.~\ref{fig:visual3} compare visual results under multiple out-of-distribution spatiotemporal scales. Even with large spatial upscaling, our method stably reconstructs frames with clear textures while maintaining excellent temporal consistency.

\textbf{Temporal Consistency Comparison.} Firstly, according to the FloLPIPS performance in Table~\ref{tab:1} and Table~\ref{tab:2}, our method exceeds existing methods by considerable margins, demonstrating its superior capability in temporal modeling. In addition, we further evaluate temporal consistency using frame-wise temporal profiles in Fig.~\ref{fig:temporal}. Existing methods often suffer from flickering, misalignment, or temporal instability. Moreover, some DM-based methods (\textit{e.g.}, with DOVE~\cite{chen2025dove}) tend to produce over-sharpened results that are less well aligned with GT details. In contrast, OSDEnhancer exhibits smoother temporal transitions and faithful textures and structures, achieving superior temporal coherence.

\textbf{Complexity Discussion.}
Table~\ref{tab:complexity} compares diffusion steps and inference time for generating a 97-frame $1024\times1024$ video on MVSR4x~\cite{wang2023benchmark} using the same NVIDIA A800 GPU. Two-stage pipelines often incur substantial latency due to multiple diffusion models with multi-step inference (\emph{e.g.}, LDMVFI~\cite{danier2024ldmvfi} + STAR~\cite{xie2025star}). Although replacing LDMVFI~\cite{danier2024ldmvfi} with EDEN~\cite{zhang2025eden}, which uses fewer diffusion steps, reduces the VFI cost, the overall runtime is still dominated by the subsequent VSR stage. VEnhancer~\cite{he2024venhancer} requires 15 diffusion steps for joint STVSR, leading to high inference latency. Benefiting from unified STVSR in one-step diffusion, OSDEnhancer involves much lower latency, demonstrating a preferable trade-off between efficiency and effectiveness.

\input{Tables/Table3}
\input{Tables/TableA1}

\subsection{Ablation Study}
To investigate the effectiveness of the proposed methodological design, we conduct ablation studies by maintaining the training protocols used in the main experiments and report PSNR for reconstruction fidelity, LPIPS~\cite{zhang2018unreasonable} for perceptual quality, and FloLPIPS~\cite{danier2022flolpips} for motion-aware temporal consistency on the UDM10 dataset~\cite{tao2017detail}.

\begin{figure}[!t]
  \includegraphics[width=\columnwidth]{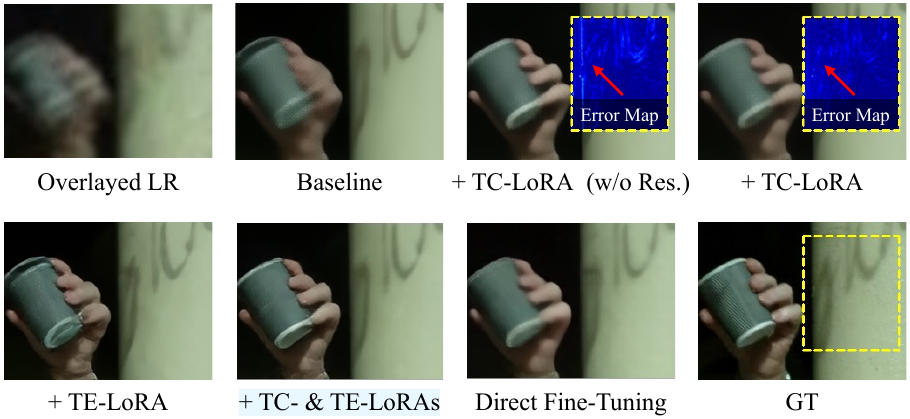}
  \caption{Visual ablation results of the divide-and-conquer adaptation scheme on UDM10~\cite{tao2017detail} under $4\times$ spatial upscaling and single-frame interpolation. Yellow boxes show error maps computed from the corresponding GT regions.}
  \label{fig:A1}
\end{figure}

\textbf{Divide-and-Conquer Adaptation.}
In OSDEnhancer, we present the divide-and-conquer adaptation approach through the TC- and TE-LoRAs. Here, we adopt the one-step adapted DiT as the baseline, along with variants adding the TC-LoRA without temporal residuals, TC-LoRA, TE-LoRA, and a direct fine-tuning approach. From Table~\ref{tab:A1}, we observe that introducing the TC-LoRA without residuals slightly improves LPIPS and FloLPIPS but leads to a PSNR drop, indicating that temporal adaptation without explicit residual cues provides insufficient motion guidance. By contrast, integrating the TC-LoRA with residual sequences to explicitly model inter-frame dynamics yields a 0.12 dB PSNR gain alongside improved LPIPS and FloLPIPS. The TE-LoRA compels the model to emphasize local textures, substantially enhancing reconstruction fidelity. The synergistic integration of both LoRAs achieves optimal performance, validating their excellent effect. Moreover, we also conduct direct fine-tuning on the baseline. Though it performs better than the individual LoRA, it is still inferior to our divide-and-conquer adaptation scheme. Fig.~\ref{fig:A1} shows a visual comparison, where the error maps further reveal that the residual-guided TC-LoRA enables more accurate synthesis by exploiting temporal coherence while maintaining static structural integrity. We can see that both the baseline and the TC-only variant produce overly smooth outputs due to compressed latent-space supervision, while the TE-only variant exhibits temporal ghosting due to the lack of temporal coherence. The superiority of the full model demonstrates that the complementary effects of the TC- and TE-LoRAs can effectively address both over-smoothing and ghosting issues, showing more promising restoration.

\input{Tables/TableA2}

\begin{figure}[t]
  \includegraphics[width=\columnwidth]{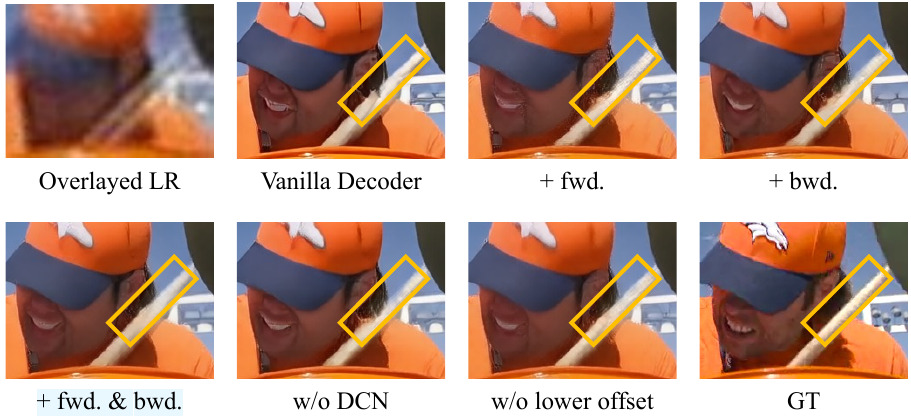}
  \caption{Visual ablation results of bidirectional deformable VAE decoder on UDM10~\cite{tao2017detail} under $4\times$ spatial upscaling and single-frame interpolation.}
  \label{fig:A2}
\end{figure}

\input{Tables/TableA3}
\input{Tables/TableA4}

\textbf{Bidirectional Deformable VAE Decoder.}
We evaluate the proposed bidirectional deformable VAE decoder by establishing the vanilla VAE decoder as a baseline and comparing it against progressive compensation variants and module ablations. Quantitatively, Table~\ref{tab:A2} demonstrates that adding forward feature propagation increases the baseline PSNR from 25.48 dB to 26.11 dB, while integrating backward propagation to leverage future frames further improves it to 26.23 dB. Besides, the model with full bidirectional compensation achieves the highest PSNR and the best LPIPS and FloLPIPS. Conversely, removing deformable convolutions and applying cross-attention directly (w/o DCN), or discarding multi-scale offset aggregation (w/o lower offset), leads to notable performance drops, confirming that both mechanisms are necessary for accurate feature alignment. Qualitatively, Fig.~\ref{fig:A2} visualizes these effects. While the vanilla baseline produces severe artifacts in interpolated frames due to missing temporal context, the complete bidirectional model effectively suppresses these errors and aligns textures closely with GT. Furthermore, visual deviations emerge when deformable convolutions or multi-scale offsets are removed, demonstrating that both designs are crucial for accurately capturing spatiotemporal dependencies.

\textbf{Loss Configuration.} Our OSDEnhancer involves different loss configurations in the TC and TE adaptations. The supervision in the TC adaptation includes an MSE loss $\mathcal{L}_\mathrm{mse}$ and a residual-aware loss $\mathcal{L}_\mathrm{res}$, with the results shown in Table~\ref{tab:A3}. We can see that using only the MSE loss $\mathcal{L}_\mathrm{mse}$ achieves the highest PSNR, while adding the residual supervision $\mathcal{L}_\mathrm{res}$ slightly reduces PSNR but improves LPIPS and FloLPIPS, indicating better perceptual quality and temporal consistency. This suggests that residual modeling additionally helps refine inter-frame details beyond strict MSE fitting.

In the TE adaptation, in addition to the metrics reported in Table~\ref{tab:A3}, we further report MUSIQ~\cite{ke2021musiq} and DOVER~\cite{wu2023exploring} in Table~\ref{tab:A4}. Using only $\mathcal{L}_1$ achieves the best PSNR but results in inferior perceptual and temporal metrics, reflecting over-smoothed outputs. Adding the textural loss $\mathcal{L}_\mathrm{dists}$ substantially improves LPIPS to 0.249 and significantly boosts MUSIQ to 58.86. The further introduction of a no-reference quality assessment loss $\mathcal{L}_\mathrm{nqa}$ brings additional gains in perceptual and video quality, raising MUSIQ and DOVER to 64.98 and 0.780, respectively. We also evaluate the optical-flow warping loss $\mathcal{L}_\mathrm{warp}$ for the temporal consistency of enhanced textures. As we can see, applying it to all frequency components offers limited benefits, whereas restricting it to high-frequency components (HF-only) achieves the best overall performance across perceptual metrics. This indicates that enforcing inter-frame consistency primarily on high-frequency textures better preserves fine structures and avoids over-constraining low-frequency regions, leading to superior performance.

\section{Conclusion}
In this paper, we present OSDEnhancer, a novel one-step diffusion framework for real-world STVSR. By proposing a progressive divide-and-conquer adaptation scheme, our method employs dedicated TC- and TE-LoRAs on a shared DiT backbone to collaboratively model inter-frame dynamics and enrich fine-grained textures. Furthermore, a bidirectional deformable VAE decoder is introduced to facilitate precise motion compensation and strengthen spatiotemporal dependencies during latent-to-pixel reconstruction. Extensive experiments demonstrate that OSDEnhancer achieves superior visual fidelity and temporal consistency across arbitrary spatiotemporal scales and complex degradations while maintaining favorable inference efficiency, highlighting its immense potential for practical applications.

\bibliographystyle{IEEEtran}
\bibliography{main}


 




\vfill

\end{document}

%% file: Tables/Table1v2.tex
\definecolor{RankA}{HTML}{FF0000} 
\definecolor{RankB}{HTML}{0000FF} 
\definecolor{RankC}{HTML}{1E5AA8} 
\definecolor{OursBG}{HTML}{E8F7FF}
\newcommand{\rankA}[1]{{\color{RankA}\textbf{#1}}} 
\newcommand{\rankB}[1]{{\color{RankB}#1}} 
\newcolumntype{C}[1]{>{\centering\arraybackslash}m{#1}}

\begin{table*}[t]
\centering
\caption{Quantitative Comparisons on Multiple Datasets including Synthetic Datasets UDM10~\cite{tao2017detail}, SPMCS~\cite{yi2019progressive}, and YouHQ40~\cite{zhou2024upscale}, and Real-World datasets MVSR4x~\cite{wang2023benchmark} and VideoLQ~\cite{chan2022investigating}. \textcolor{RankA}{Red} and \textcolor{RankB}{blue} Indicate the Best and Second-best, Respectively.}
\label{tab:1}
\resizebox{\textwidth}{!}{
\begin{tabular}{c|c|*{3}{C{1cm}}|*{3}{C{1cm}}|c*{1}{C{1cm}}c*{2}{C{1cm}}|cc}
\hline
\multirow{3}{*}{Datasets} & \multirow{3}{*}{Metrics}
& \multicolumn{3}{c|}{LDMVFI~\cite{danier2024ldmvfi}}
& \multicolumn{3}{c|}{EDEN~\cite{zhang2025eden}}
& \multirow{3}{*}{\shortstack{VideoINR\\\cite{chen2022videoinr}}}
& \multirow{3}{*}{\shortstack{MoTIF\\\cite{chen2023motif}}} 
& \multirow{3}{*}{\shortstack{BF-STVSR\\\cite{kim2025bf}}} 
& \multirow{3}{*}{\shortstack{STNO\\\cite{zhang2025space}}} 
& \multirow{3}{*}{\shortstack{V$^3$\\\cite{becker2025continuous}}} 
& \multirow{3}{*}{\shortstack{VEnhancer\\\cite{he2024venhancer}}} 
&\cellcolor{OursBG} \\
\cline{3-8}
& 
&  \multirow{2}{*}{\shortstack{STAR\\\cite{xie2025star}}} & \multirow{2}{*}{\shortstack{DOVE\\\cite{chen2025dove}}} & \multirow{2}{*}{\shortstack{SeedVR2\\\cite{wang2025seedvr2}}}
&  \multirow{2}{*}{\shortstack{STAR\\\cite{xie2025star}}} & \multirow{2}{*}{\shortstack{DOVE\\\cite{chen2025dove}}} & \multirow{2}{*}{\shortstack{SeedVR2\\\cite{wang2025seedvr2}}}
&  &  &  &  &  &  &\cellcolor{OursBG} \\
& 
& & & 
& & & 
&  &  &  &  &  &  &\cellcolor{OursBG}\multirow{-3}{*}{\shortstack{\textbf{OSDEnhancer}\\\textbf{(Ours)}}} \\
\hline

\multirow{9}{*}{UDM10}
& PSNR~$\uparrow$ & 24.04 & \rankB{26.37} & 25.52 & 23.97 & 26.36 & 25.55 & 25.27 & 24.91 & 25.07 & 25.42 & 24.80 & 21.64 & \rankA{26.44}\cellcolor{OursBG} \\
& SSIM~$\uparrow$ & 0.686 & \rankB{0.766} & 0.731 & 0.686 & 0.765 & 0.734 & 0.726 & 0.708 & 0.716 & 0.734 & 0.649 & 0.679 & \rankA{0.775}\cellcolor{OursBG} \\
& LPIPS~$\downarrow$ & 0.429 & 0.279 & 0.289 & 0.426 & 0.279 & \rankB{0.278} & 0.364 & 0.396 & 0.377 & 0.352 & 0.529 & 0.451 & \rankA{0.248}\cellcolor{OursBG} \\
& DISTS~$\downarrow$ & 0.225 & 0.155 & 0.146 & 0.224 & 0.156 & \rankB{0.141} & 0.233 & 0.240 & 0.243 & 0.230 & 0.401 & 0.257 & \rankA{0.118}\cellcolor{OursBG} \\
& FloLPIPS~$\downarrow$ & 0.426 & 0.282 & 0.287 & 0.426 & 0.284 & \rankB{0.280} & 0.363 & 0.400 & 0.379 & 0.354 & 0.529 & 0.428 & \rankA{0.253}\cellcolor{OursBG} \\
& MUSIQ~$\uparrow$ & 36.60 & 60.70 & 50.55 & 36.26 & \rankB{60.87} & 50.56 & 43.72 & 43.91 & 42.83 & 43.31 & 22.95 & 33.55 & \rankA{65.95}\cellcolor{OursBG} \\
& CLIP-IQA~$\uparrow$ & 0.248 & \rankB{0.463} & 0.392 & 0.247 & 0.462 & 0.387 & 0.294 & 0.324 & 0.301 & 0.325 & 0.191 & 0.272 & \rankA{0.490}\cellcolor{OursBG} \\
& FasterVQA~$\uparrow$ & 0.591 & \rankB{0.764} & 0.621 & 0.597 & 0.760 & 0.613 & 0.639 & 0.731 & 0.700 & 0.666 & 0.178 & 0.524 & \rankA{0.805}\cellcolor{OursBG} \\
& DOVER~$\uparrow$ & 0.441 & 0.770 & 0.520 & 0.466 & \rankB{0.776} & 0.529 & 0.422 & 0.547 & 0.499 & 0.529 & 0.108 & 0.447 & \rankA{0.796}\cellcolor{OursBG} \\

\hline
\multirow{9}{*}{SPMCS}
& PSNR~$\uparrow$ & 21.31 & 22.98 & 22.47 & 21.30 & 22.97 & 22.43 & 22.91 & 22.89 & 22.90 & \rankB{23.17} & 22.88 & 19.40 & \rankA{23.27}\cellcolor{OursBG} \\
& SSIM~$\uparrow$ & 0.537 & \rankB{0.614} & 0.603 & 0.538 & \rankB{0.614} & 0.603 & 0.594 & 0.596 & 0.591 & 0.606 & 0.567 & 0.502 & \rankA{0.617}\cellcolor{OursBG} \\
& LPIPS~$\downarrow$ & 0.555 & 0.293 & \rankB{0.272} & 0.550 & 0.294 & \rankA{0.270} & 0.400 & 0.388 & 0.391 & 0.365 & 0.521 & 0.533 & 0.288\cellcolor{OursBG} \\
& DISTS~$\downarrow$ & 0.296 & 0.172 & 0.162 & 0.292 & 0.173 & \rankB{0.161} & 0.280 & 0.278 & 0.281 & 0.262 & 0.374 & 0.268 & \rankA{0.152}\cellcolor{OursBG} \\
& FloLPIPS~$\downarrow$ & 0.514 & 0.273 & \rankB{0.251} & 0.515 & 0.275 & \rankA{0.249} & 0.383 & 0.369 & 0.376 & 0.343 & 0.482 & 0.510 & 0.270\cellcolor{OursBG} \\
& MUSIQ~$\uparrow$ & 34.79 & \rankB{69.18} & 65.65 & 35.56 & 69.15 & 65.39 & 44.80 & 48.42 & 48.81 & 50.66 & 28.53 & 41.67 & \rankA{72.61}\cellcolor{OursBG} \\
& CLIP-IQA~$\uparrow$ & 0.265 & 0.519 & \rankB{0.528} & 0.262 & 0.519 & 0.522 & 0.267 & 0.305 & 0.285 & 0.333 & 0.183 & 0.326 & \rankA{0.535}\cellcolor{OursBG} \\
& FasterVQA~$\uparrow$ & 0.391 & \rankB{0.720} & 0.684 & 0.423 & \rankB{0.720} & 0.687 & 0.575 & 0.587 & 0.583 & 0.655 & 0.201 & 0.454 & \rankA{0.777}\cellcolor{OursBG} \\
& DOVER~$\uparrow$ & 0.301 & 0.778 & 0.659 & 0.308 & \rankB{0.781} & 0.657 & 0.412 & 0.445 & 0.458 & 0.470 & 0.123 & 0.399 & \rankA{0.792}\cellcolor{OursBG} \\

\hline
\multirow{9}{*}{YouHQ40}
& PSNR~$\uparrow$ & 22.74 & \rankB{24.22} & 23.39 & 22.74 & 24.21 & 23.37 & 23.92 & 23.75 & 23.85 & 24.01 & 23.76 & 20.63 & \rankA{24.24}\cellcolor{OursBG} \\
& SSIM~$\uparrow$ & 0.641 & \rankB{0.675} & 0.661 & 0.640 & \rankB{0.675} & 0.660 & 0.655 & 0.644 & 0.649 & 0.658 & 0.603 & 0.600 & \rankA{0.678}\cellcolor{OursBG} \\
& LPIPS~$\downarrow$ & 0.467 & 0.300 & \rankA{0.256} & 0.469 & 0.300 & \rankB{0.257} & 0.390 & 0.375 & 0.373 & 0.369 & 0.540 & 0.487 & 0.287\cellcolor{OursBG} \\
& DISTS~$\downarrow$ & 0.225 & 0.149 & \rankA{0.118} & 0.225 & 0.149 & \rankB{0.119} & 0.206 & 0.205 & 0.205 & 0.200 & 0.372 & 0.224 & 0.120\cellcolor{OursBG} \\
& FloLPIPS~$\downarrow$ & 0.445 & 0.304 & \rankA{0.254} & 0.446 & 0.304 & \rankB{0.257} & 0.392 & 0.375 & 0.379 & 0.365 & 0.532 & 0.463 & 0.289\cellcolor{OursBG} \\
& MUSIQ~$\uparrow$ & 33.63 & 60.82 & \rankB{64.25} & 33.64 & 60.88 & 64.13 & 40.33 & 46.11 & 45.27 & 41.47 & 24.87 & 40.06 & \rankA{65.66}\cellcolor{OursBG} \\
& CLIP-IQA~$\uparrow$ & 0.275 & 0.446 & \rankA{0.499} & 0.275 & 0.447 & \rankB{0.497} & 0.295 & 0.360 & 0.333 & 0.342 & 0.223 & 0.333 & 0.495\cellcolor{OursBG} \\
& FasterVQA~$\uparrow$ & 0.568 & 0.857 & \rankB{0.873} & 0.565 & 0.856 & 0.871 & 0.749 & 0.801 & 0.790 & 0.786 & 0.348 & 0.688 & \rankA{0.886}\cellcolor{OursBG} \\
& DOVER~$\uparrow$ & 0.580 & 0.851 & \rankA{0.871} & 0.580 & 0.850 & \rankA{0.871} & 0.671 & 0.725 & 0.744 & 0.709 & 0.403 & 0.675 & \rankB{0.867}\cellcolor{OursBG} \\

\hline
\hline
\multirow{9}{*}{MVSR4x}
& PSNR~$\uparrow$ & 22.52 & 22.30 & \rankA{23.03} & 22.51 & 22.28 & \rankB{23.01} & 21.67 & 19.91 & 21.42 & 22.34 & 22.69 & 20.87 & 22.71\cellcolor{OursBG} \\
& SSIM~$\uparrow$ & 0.748 & 0.751 & \rankB{0.765} & 0.748 & 0.750 & 0.763 & 0.736 & 0.734 & 0.749 & 0.731 & \rankA{0.766} & 0.734 & \rankB{0.765}\cellcolor{OursBG} \\
& LPIPS~$\downarrow$ & 0.410 & \rankB{0.348} & 0.349 & 0.411 & 0.349 & 0.357 & 0.459 & 0.497 & 0.485 & 0.464 & 0.431 & 0.440 & \rankA{0.342}\cellcolor{OursBG} \\
& DISTS~$\downarrow$ & 0.257 & 0.237 & \rankB{0.226} & 0.259 & 0.238 & 0.232 & 0.305 & 0.286 & 0.336 & 0.324 & 0.289 & 0.276 & \rankA{0.223}\cellcolor{OursBG} \\
& FloLPIPS~$\downarrow$ & 0.409 & \rankB{0.345} & 0.346 & 0.411 & 0.346 & 0.351 & 0.460 & 0.497 & 0.478 & 0.436 & 0.418 & 0.399 & \rankA{0.344}\cellcolor{OursBG} \\
& MUSIQ~$\uparrow$ & 29.70 & \rankA{62.69} & 31.35 & 30.06 & 62.58 & 30.11 & 34.25 & 26.10 & 21.07 & 41.37 & 24.09 & 35.22 & \rankB{62.65}\cellcolor{OursBG} \\
& CLIP-IQA~$\uparrow$ & 0.257 & \rankB{0.521} & 0.204 & 0.256 & \rankA{0.524} & 0.196 & 0.367 & 0.260 & 0.315 & 0.479 & 0.286 & 0.318 & 0.514\cellcolor{OursBG} \\
& FasterVQA~$\uparrow$ & 0.264 & \rankB{0.775} & 0.245 & 0.279 & \rankB{0.775} & 0.202 & 0.558 & 0.304 & 0.245 & 0.760 & 0.138 & 0.332 & \rankA{0.778}\cellcolor{OursBG} \\
& DOVER~$\uparrow$ & 0.204 & \rankA{0.706} & 0.204 & 0.202 & \rankB{0.702} & 0.185 & 0.258 & 0.138 & 0.142 & 0.432 & 0.137 & 0.276 & 0.665\cellcolor{OursBG} \\

\hline
\multirow{4}{*}{VideoLQ}
& MUSIQ~$\uparrow$ & 39.14 & 43.84 & 36.29 & 39.17 & \rankB{43.93} & 35.98 & 34.20 & 36.66 & 37.29 & 31.63 & 22.80 & 39.38 & \rankA{45.30}\cellcolor{OursBG} \\
& CLIP-IQA~$\uparrow$ & 0.289 & 0.287 & 0.227 & 0.290 & 0.286 & 0.225 & 0.246 & 0.275 & 0.247 & 0.228 & 0.236 & \rankB{0.303} & \rankA{0.351}\cellcolor{OursBG} \\
& FasterVQA~$\uparrow$ & 0.663 & 0.718 & 0.594 & 0.673 & \rankB{0.721} & 0.593 & 0.619 & 0.667 & 0.675 & 0.604 & 0.324 & 0.639 & \rankA{0.765}\cellcolor{OursBG} \\
& DOVER~$\uparrow$ & 0.712 & 0.748 & 0.664 & 0.712 & \rankA{0.751} & 0.661 & 0.643 & 0.675 & 0.668 & 0.603 & 0.476 & 0.665 & \rankB{0.749}\cellcolor{OursBG} \\

\hline
\end{tabular}
}
\end{table*}

%% file: Tables/Table2v2.tex
\begin{table*}[htbp]
\centering
\caption{Quantitative Comparisons (PSNR $\uparrow$ / LPIPS $\downarrow$ / FloLPIPS $\downarrow$) with State-of-the-Art Continuous STVSR Methods under Different Temporal and Spatial Scales on the GoPro Dataset~\cite{nah2017deep}. \textcolor{RankA}{Red} Indicates the Best Performance.}
\label{tab:2}
\resizebox{\textwidth}{!}{
\begin{tabular}{c|c|ccccc|cc}
\hline
Temporal & Spatial
& \multirow{2}{*}{VideoINR~\cite{chen2022videoinr}}
& \multirow{2}{*}{MoTIF~\cite{chen2023motif}}
& \multirow{2}{*}{BF-STVSR~\cite{kim2025bf}}
& \multirow{2}{*}{STNO~\cite{zhang2025space}}
& \multirow{2}{*}{V$^3$~\cite{becker2025continuous}}
& \multirow{2}{*}{VEnhancer~\cite{he2024venhancer}}
& \cellcolor{OursBG} \\
Scale & Scale & & & & & & & \multirow{-2}{*}{\textbf{OSDEnhancer (Ours)}}\cellcolor{OursBG} \\
\hline
\makecell{$8\times$\\($K=7$)} & $4\times$
& 24.10 / 0.305 / 0.300
& 23.99 / 0.297 / 0.295
& 24.22 / 0.276 / 0.274
& 23.42 / 0.308 / 0.315
& 24.19 / 0.494 / 0.490
& 19.74 / 0.541 / 0.535
& \rankA{24.53} / \rankA{0.269} / \rankA{0.263}\cellcolor{OursBG} \\
\hline
\hline
\multirow{3}{*}{\shortstack{$1\times$\\($K=0$)}} 
& $8\times$ 
& 23.42 / 0.436 / 0.426 
& 23.45 / 0.418 / 0.411 
& 23.49 / 0.448 / 0.437 
& 23.60 / 0.424 / 0.410 
& 23.59 / 0.576 / 0.569 
& 21.56 / 0.545 / 0.532 
& \rankA{24.05} / \rankA{0.302} / \rankA{0.295}\cellcolor{OursBG} \\
& $12\times$ 
& 22.28 / 0.520 / 0.513 
& 22.20 / 0.507 / 0.501 
& 22.16 / 0.535 / 0.528 
& 22.31 / 0.512 / 0.503 
& 22.48 / 0.651 / 0.643 
& 20.12 / 0.622 / 0.613 
& \rankA{22.97} / \rankA{0.353} / \rankA{0.341}\cellcolor{OursBG} \\
& $16\times$ 
& 21.17 / 0.598 / 0.591 
& 20.98 / 0.566 / 0.562 
& 21.08 / 0.555 / 0.552 
& 21.33 / 0.586 / 0.577 
& 21.49 / 0.680 / 0.673 
& 19.78 / 0.690 / 0.675 
& \rankA{21.95} / \rankA{0.412} / \rankA{0.396}\cellcolor{OursBG} \\
\hline
\multirow{3}{*}{\shortstack{$6\times$\\($K=5$)}} 
& $8\times$ 
& 23.27 / 0.434 / 0.435 
& 23.36 / 0.409 / 0.403 
& 23.45 / 0.430 / 0.427 
& 23.34 / 0.349 / 0.351 
& 23.45 / 0.571 / 0.567 
& 14.09 / 0.716 / 0.717 
& \rankA{23.55} / \rankA{0.321} / \rankA{0.314}\cellcolor{OursBG} \\
& $12\times$ 
& 22.19 / 0.516 / 0.509 
& 22.25 / 0.499 / 0.487 
& 22.26 / 0.525 / 0.517 
& 22.40 / 0.452 / 0.450 
& 22.45 / 0.622 / 0.619 
& 14.20 / 0.748 / 0.749 
& \rankA{22.63} / \rankA{0.375} / \rankA{0.363}\cellcolor{OursBG} \\
& $16\times$ 
& 21.13 / 0.592 / 0.583 
& 21.02 / 0.567 / 0.558 
& 21.20 / 0.557 / 0.550 
& 21.56 / 0.546 / 0.540 
& 21.57 / 0.653 / 0.648 
& 14.53 / 0.773 / 0.773 
& \rankA{21.75} / \rankA{0.432} / \rankA{0.417}\cellcolor{OursBG} \\
\hline
\multirow{3}{*}{\shortstack{$12\times$\\($K=11$)}} 
& $8\times$ 
& 22.52 / 0.431 / 0.432 
& 22.55 / 0.418 / 0.421 
& \rankA{22.72} / 0.448 / 0.448 
& 21.82 / 0.381 / 0.392 
& 22.54 / 0.584 / 0.583 
& 13.68 / 0.737 / 0.738 
& 22.25 / \rankA{0.350} / \rankA{0.347}\cellcolor{OursBG} \\
& $12\times$ 
& 21.67 / 0.512 / 0.506 
& 21.70 / 0.504 / 0.498 
& 21.81 / 0.536 / 0.531 
& 21.24 / 0.473 / 0.480 
& \rankA{21.82} / 0.630 / 0.630 
& 13.72 / 0.759 / 0.760 
& 21.62 / \rankA{0.399} / \rankA{0.392}\cellcolor{OursBG} \\
& $16\times$ 
& 20.81 / 0.587 / 0.580 
& 20.72 / 0.569 / 0.563 
& 20.95 / 0.573 / 0.567 
& 20.81 / 0.556 / 0.559 
& \rankA{21.17} / 0.657 / 0.654 
& 14.07 / 0.784 / 0.784 
& 21.05 / \rankA{0.450} / \rankA{0.440}\cellcolor{OursBG} \\
\hline
\end{tabular}}
\end{table*}

%% file: Tables/Table3.tex
\begin{table}[t]
\centering
\caption{Complexity comparison among DM-based methods. All Methods Are Evaluated on the same NVIDIA A800 GPU by Generating a 97-frame $1024\times1024$ Video with Single-Frame Interpolation on MVSR4x~\cite{wang2023benchmark}.}
\label{tab:complexity}
\setlength{\tabcolsep}{3pt}    
\renewcommand{\arraystretch}{1}
\scriptsize         
\begin{tabular}{l | ll}
\hline
Method & Diffusion Step & Inference Time (s)\\
\hline
LDMVFI~\cite{danier2024ldmvfi} + STAR~\cite{xie2025star}    & 200 + 15 & 854 (348 + 506) \\
LDMVFI~\cite{danier2024ldmvfi} + DOVE~\cite{chen2025dove}    & 200 + 1 & 414 (348 + 66) \\
LDMVFI~\cite{danier2024ldmvfi} + SeedVR2~\cite{wang2025seedvr2}    & 200 + 1 & 488 (348 + 140) \\
\hline
EDEN~\cite{zhang2025eden} + STAR~\cite{xie2025star}   & 2 + 15 & 516 (10 + 506) \\
EDEN~\cite{zhang2025eden} + DOVE~\cite{chen2025dove}    & 2 + 1 & 76 (10 + 66) \\
EDEN~\cite{zhang2025eden} + SeedVR2~\cite{wang2025seedvr2}    & 2 + 1 & 150 (10 + 140) \\
\hline
VEnhancer~\cite{he2024venhancer}      & 15 & 871 \\
\rowcolor{OursBG}\textbf{OSDEnhancer (Ours)} & 1 & 129 \\
\hline
\end{tabular}
\end{table}

%% file: Tables/TableA1.tex
\begin{table}[!t]
\centering
\caption{Ablation Study on the Divide-and-Conquer Adaptation through TC- and TE-LoRAs.}
\label{tab:A1}
\setlength{\tabcolsep}{3pt}     
\renewcommand{\arraystretch}{1}
\scriptsize                   
\begin{tabular}{l| c c c}
\hline
 & PSNR $\uparrow$ & LPIPS $\downarrow$  & FloLPIPS $\downarrow$ \\
\hline
Baseline & 26.77 & 0.324 & 0.320 \\
+ TC-LoRA (w/o Residuals) & 26.51 & 0.323 & 0.313 \\
+ TC-LoRA & \textbf{26.89} & 0.320 & 0.313 \\
+ TE-LoRA & 25.98 & 0.251 & 0.257 \\
\rowcolor{OursBG}+ TC- \& TE-LoRAs & 26.44 & \textbf{0.248} & \textbf{0.253} \\
\hline
Direct Fine-Tuning & 26.17 & 0.301 & 0.307 \\
\hline
\end{tabular}
\end{table}

%% file: Tables/TableA2.tex
\begin{table}[!t]
\centering
\caption{Ablation Study on the Bidirectional Deformable VAE Decoder.
}
\label{tab:A2}
\setlength{\tabcolsep}{3pt}
\renewcommand{\arraystretch}{1}
\footnotesize 
\resizebox{\columnwidth}{!}{
\begin{tabular}{l | c c c}
\hline
 & PSNR $\uparrow$ & LPIPS $\downarrow$  & FloLPIPS $\downarrow$ \\
\hline
Vanilla VAE Decoder & 25.48 & 0.257 & 0.260 \\
+ fwd. Compensation & 26.11 & 0.253 & \textbf{0.253} \\
+ bwd. Compensation & 26.23 & 0.253 & \textbf{0.253} \\
\rowcolor{OursBG}+ fwd. \& bwd. Compensation & \textbf{26.44} & \textbf{0.248} & \textbf{0.253} \\
+ fwd. \& bwd. Compensation (w/o DCN) & 25.98 & 0.256 & 0.258 \\
+ fwd. \& bwd. Compensation (w/o lower offset) & 26.31 & 0.254 & 0.256 \\
\hline
\end{tabular}
}
\end{table}

%% file: Tables/TableA3.tex
\begin{table}[t]
\centering
\caption{Ablation Study on Loss Configurations in the TC Adaptation.}
\label{tab:A3}
\setlength{\tabcolsep}{3pt}
\renewcommand{\arraystretch}{1}
\scriptsize
\begin{tabular}{c c | c c c}
\hline
$\mathcal{L}_\mathrm{mse}$ & $\mathcal{L}_\mathrm{res}$ & PSNR $\uparrow$ & LPIPS $\downarrow$ & FloLPIPS $\downarrow$\\
\hline
\checkmark &  & \textbf{26.91} & 0.324 & 0.317 \\
\rowcolor{OursBG}\checkmark & \checkmark &
26.89 & \textbf{0.320} & \textbf{0.313} \\
\hline
\end{tabular}
\end{table}

%% file: Tables/TableA4.tex
\begin{table}[!t]
\centering
\caption{Ablation study on Loss Configurations in the TE Adaptation. All: All Frequency Components Are Involved in Optical-Flow Warp Loss; HF-only: Only High-Frequency Components Are Involved.}
\label{tab:A4}
\setlength{\tabcolsep}{3pt}
\renewcommand{\arraystretch}{1}
\footnotesize
\resizebox{\columnwidth}{!}{
\begin{tabular}{c c c c | c c c c c}
\hline
$\mathcal{L}_\mathrm{1}$ & $\mathcal{L}_\mathrm{dists}$ & $\mathcal{L}_\mathrm{nqa}$ & $\mathcal{L}_\mathrm{warp}$ & PSNR $\uparrow$ & LPIPS $\downarrow$ & FloLPIPS $\downarrow$ & MUSIQ $\uparrow$ & DOVER $\uparrow$\\
\hline
\checkmark &  &  &  &
\textbf{27.23} & 0.310 & 0.294 & 47.55 & 0.572 \\
\checkmark & \checkmark &  &  &
26.40 & 0.249 & 0.256 & 58.86 & 0.727  \\
\checkmark & \checkmark & \checkmark &  &
26.45 & 0.255 & 0.260 & 64.98 & 0.780  \\
\checkmark & \checkmark & \checkmark & All &
26.40 & 0.252 & 0.254 & 64.84 & 0.784  \\
\rowcolor{OursBG} \checkmark & \checkmark & \checkmark & HF-only  &
26.44 & \textbf{0.248} & \textbf{0.253} & \textbf{65.95} & \textbf{0.796} \\
\hline
\end{tabular}}
\end{table}

%% file: main.bib
@inproceedings{tian2021self,
  title={Self-conditioned probabilistic learning of video rescaling},
  author={Tian, Yuan and Lu, Guo and Min, Xiongkuo and Che, Zhaohui and Zhai, Guangtao and Guo, Guodong and Gao, Zhiyong},
  booktitle={Proceedings of the IEEE/CVF International Conference on Computer Vision},
  pages={4490--4499},
  year={2021}
}

@article{qiu2023learning,
  author={Qiu, Zhongwei and Yang, Huan and Fu, Jianlong and Liu, Daochang and Xu, Chang and Fu, Dongmei},
  journal={IEEE Transactions on Pattern Analysis and Machine Intelligence}, 
  title={Learning degradation-robust spatiotemporal frequency-transformer for video super-resolution}, 
  year={2023},
  volume={45},
  number={12},
  pages={14888-14904}
}

@article{li2024enhanced,
  title={Enhanced video super-resolution network towards compressed data},
  author={Li, Feng and Wu, Yixuan and Li, Anqi and Bai, Huihui and Cong, Runmin and Zhao, Yao},
  journal={ACM Transactions on Multimedia Computing, Communications and Applications},
  volume={20},
  number={7},
  pages={1--21},
  year={2024}
}

@inproceedings{bao2019depth,
  title={Depth-aware video frame interpolation},
  author={Bao, Wenbo and Lai, Wei-Sheng and Ma, Chao and Zhang, Xiaoyun and Gao, Zhiyong and Yang, Ming-Hsuan},
  booktitle={Proceedings of the IEEE/CVF Conference on Computer Vision and Pattern Recognition},
  pages={3703--3712},
  year={2019}
}

@article{shen2020video,
  title={Video frame interpolation and enhancement via pyramid recurrent framework},
  author={Shen, Wang and Bao, Wenbo and Zhai, Guangtao and Chen, Li and Min, Xiongkuo and Gao, Zhiyong},
  journal={IEEE Transactions on Image Processing},
  volume={30},
  pages={277--292},
  year={2020}
}

@inproceedings{xiang2020zooming,
  title={Zooming slow-mo: Fast and accurate one-stage space-time video super-resolution},
  author={Xiang, Xiaoyu and Tian, Yapeng and Zhang, Yulun and Fu, Yun and Allebach, Jan P and Xu, Chenliang},
  booktitle={Proceedings of the IEEE/CVF Conference on Computer Vision and Pattern Recognition},
  pages={3367--3376},
  year={2020}
}

@inproceedings{haris2020space,
  title={Space-time-aware multi-resolution video enhancement},
  author={Haris, Muhammad and Shakhnarovich, Greg and Ukita, Norimichi},
  booktitle={Proceedings of the IEEE/CVF Conference on Computer Vision and Pattern Recognition},
  pages={2859--2868},
  year={2020}
}

@inproceedings{xiao2020space,
  title={Space-time video super-resolution using temporal profiles},
  author={Xiao, Zeyu and Xiong, Zhiwei and Fu, Xueyang and Liu, Dong and Zha, Zheng-Jun},
  booktitle={Proceedings of the 28th ACM International Conference on Multimedia},
  pages={664--672},
  year={2020}
}

@inproceedings{xu2021temporal,
  title={Temporal modulation network for controllable space-time video super-resolution},
  author={Xu, Gang and Xu, Jun and Li, Zhen and Wang, Liang and Sun, Xing and Cheng, Ming-Ming},
  booktitle={Proceedings of the IEEE/CVF Conference on Computer Vision and Pattern Recognition},
  pages={6384--6393},
  year={2021}
}

@inproceedings{geng2022rstt,
  title={Rstt: Real-time spatial temporal transformer for space-time video super-resolution},
  author={Geng, Zhicheng and Liang, Luming and Ding, Tianyu and Zharkov, Ilya},
  booktitle={Proceedings of the IEEE/CVF Conference on Computer Vision and Pattern Recognition},
  pages={17420--17430},
  year={2022}
}

@inproceedings{chen2022videoinr,
  title={Videoinr: Learning video implicit neural representation for continuous space-time super-resolution},
  author={Chen, Zeyuan and Chen, Yinbo and Liu, Jingwen and Xu, Xingqian and Goel, Vidit and Wang, Zhangyang and Shi, Humphrey and Wang, Xiaolong},
  booktitle={Proceedings of the IEEE/CVF Conference on Computer Vision and Pattern Recognition},
  pages={2037--2047},
  year={2022}
}

@inproceedings{chen2023motif,
  title={Motif: Learning motion trajectories with local implicit neural functions for continuous space-time video super-resolution},
  author={Chen, Yi-Hsin and Chen, Si-Cun and Chen, Yi-Hsin and Lin, Yen-Yu and Peng, Wen-Hsiao},
  booktitle={Proceedings of the IEEE/CVF International Conference on Computer Vision},
  pages={23074--23084},
  year={2023}
}

@inproceedings{hu2023store,
  title={Store and fetch immediately: Everything is all you need for space-time video super-resolution},
  author={Hu, Mengshun and Jiang, Kui and Nie, Zhixiang and Zhou, Jiahuan and Wang, Zheng},
  booktitle={Proceedings of the AAAI Conference on Artificial Intelligence},
  volume={37},
  number={1},
  pages={863--871},
  year={2023}
}

@inproceedings{kim2025bf,
  title={BF-STVSR: B-Splines and Fourier---Best Friends for High Fidelity Spatial-Temporal Video Super-Resolution},
  author={Kim, Eunjin and Kim, Hyeonjin and Jin, Kyong Hwan and Yoo, Jaejun},
  booktitle={Proceedings of the IEEE/CVF Conference on Computer Vision and Pattern Recognition},
  pages={28009--28018},
  year={2025}
}

@inproceedings{wei2025evenhancer,
  title={EvEnhancer: Empowering Effectiveness, Efficiency and Generalizability for Continuous Space-Time Video Super-Resolution with Events},
  author={Wei, Shuoyan and Li, Feng and Tang, Shengeng and Zhao, Yao and Bai, Huihui},
  booktitle={Proceedings of the IEEE/CVF Conference on Computer Vision and Pattern Recognition},
  pages={17755--17766},
  year={2025}
}

@article{zhang2025space,
  title={Space-time video super-resolution with neural operator},
  author={Zhang, Yuantong and Zheng, Hanyou and Yang, Daiqin and Chen, Zhenzhong and Ma, Haichuan and Ding, Wenpeng},
  journal={IEEE Transactions on Image Processing},
  volume={34},
  pages={6742-6754},
  year={2025},
  publisher={IEEE}
}

@inproceedings{chan2022investigating,
  title={Investigating tradeoffs in real-world video super-resolution},
  author={Chan, Kelvin CK and Zhou, Shangchen and Xu, Xiangyu and Loy, Chen Change},
  booktitle={Proceedings of the IEEE/CVF Conference on Computer Vision and Pattern Recognition},
  pages={5962--5971},
  year={2022}
}

@inproceedings{hu2022spatial,
  title={Spatial-temporal space hand-in-hand: Spatial-temporal video super-resolution via cycle-projected mutual learning},
  author={Hu, Mengshun and Jiang, Kui and Liao, Liang and Xiao, Jing and Jiang, Junjun and Wang, Zheng},
  booktitle={Proceedings of the IEEE/CVF Conference on Computer Vision and Pattern Recognition},
  pages={3564--3573},
  year={2022}
}

@article{hu2023cycmunet+,
  title={CycMuNet+: Cycle-projected mutual learning for spatial-temporal video super-resolution},
  author={Hu, Mengshun and Jiang, Kui and Wang, Zheng and Bai, Xiang and Hu, Ruimin},
  journal={IEEE Transactions on Pattern Analysis and Machine Intelligence},
  volume={45},
  number={11},
  pages={13376--13392},
  year={2023}
}

@inproceedings{shechtman2002increasing,
  title={Increasing space-time resolution in video},
  author={Shechtman, Eli and Caspi, Yaron and Irani, Michal},
  booktitle={Proceedings of the European Conference on Computer Vision},
  pages={753--768},
  year={2002}
}

@inproceedings{tian2020temporally,
  title={Tdan: Temporally-deformable alignment network for video super-resolution},
  author={Tian, Yapeng and Zhang, Yulun and Fu, Yun and Xu, Chenliang},
  booktitle={Proceedings of the IEEE/CVF Conference on Computer Vision and Pattern Recognition},
  pages={3357--3366},
  year={2020}
}

@inproceedings{kim2020fisr,
  title={Fisr: Deep joint frame interpolation and super-resolution with a multi-scale temporal loss},
  author={Kim, Soo Ye and Oh, Jihyong and Kim, Munchurl},
  booktitle={Proceedings of the AAAI Conference on Artificial Intelligence},
  pages={11278--11286},
  year={2020}
}

@inproceedings{rombach2022high,
  title={High-resolution image synthesis with latent diffusion models},
  author={Rombach, Robin and Blattmann, Andreas and Lorenz, Dominik and Esser, Patrick and Ommer, Björn},
  booktitle={Proceedings of the IEEE/CVF Conference on Computer Vision and Pattern Recognition},
  pages={10674--10685},
  year={2022}
}

@inproceedings{esser2024scaling,
  title={Scaling rectified flow transformers for high-resolution image synthesis},
  author={Esser, Patrick and Kulal, Sumith and Blattmann, Andreas and Entezari, Rahim and M{\"u}ller, Jonas and Saini, Harry and Levi, Yam and Lorenz, Dominik and Sauer, Axel and Boesel, Frederic and others},
  booktitle={Proceedings of the 41st International Conference on Machine Learning},
  year={2024}
}

@inproceedings{zhou2024upscale,
  title={Upscale-a-video: Temporal-consistent diffusion model for real-world video super-resolution},
  author={Zhou, Shangchen and Yang, Peiqing and Wang, Jianyi and Luo, Yihang and Loy, Chen Change},
  booktitle={Proceedings of the IEEE/CVF Conference on Computer Vision and Pattern Recognition},
  pages={2535--2545},
  year={2024}
}

@inproceedings{zhang2024realviformer,
  title={Realviformer: Investigating attention for real-world video super-resolution},
  author={Zhang, Yuehan and Yao, Angela},
  booktitle={Proceedings of the European Conference on Computer Vision},
  pages={412--428},
  year={2024},
  organization={Springer}
}

@inproceedings{xie2025star,
    author={Xie, Rui and Liu, Yinhong and Zhou, Penghao and Zhao, Chen and Zhou, Jun and Zhang, Kai and Zhang, Zhenyu and Yang, Jian and Yang, Zhenheng and Tai, Ying},
    title={STAR: Spatial-Temporal Augmentation with Text-to-Video Models for Real-World Video Super-Resolution},
    booktitle={Proceedings of the IEEE/CVF International Conference on Computer Vision},
    year={2025},
    pages={17108--17118}
}

@inproceedings{wang2025seedvr,
  title={Seedvr: Seeding infinity in diffusion transformer towards generic video restoration},
  author={Wang, Jianyi and Lin, Zhijie and Wei, Meng and Zhao, Yang and Yang, Ceyuan and Loy, Chen Change and Jiang, Lu},
  booktitle={Proceedings of the IEEE/CVF Conference on Computer Vision and Pattern Recognition},
  pages={2161--2172},
  year={2025}
}

@inproceedings{wang2025seedvr2,
  title={Seedvr2: One-step video restoration via diffusion adversarial post-training},
  author={Wang, Jianyi and Lin, Shanchuan and Lin, Zhijie and Ren, Yuxi and Wei, Meng and Yue, Zongsheng and Zhou, Shangchen and Chen, Hao and Zhao, Yang and Yang, Ceyuan and others},
  booktitle={International Conference on Learning Representations},
  year={2026}
}

@inproceedings{peng2025mitigating,
  title={Mitigating Delivery Artifacts in Real-World Video Super-Resolution},
  author={Peng, Jiaxin and Zhou, Siwang and Li, Chengqing and Li, Yucheng and Chen, Dunyun},
  booktitle={Proceedings of the 33rd ACM International Conference on Multimedia},
  pages={3114--3123},
  year={2025}
}

@inproceedings{liu2025ultravsr,
  title={Ultravsr: Achieving ultra-realistic video super-resolution with efficient one-step diffusion space},
  author={Liu, Yong and Pan, Jinshan and Li, Yinchuan and Dong, Qingji and Zhu, Chao and Guo, Yu and Wang, Fei},
  booktitle={Proceedings of the 33rd ACM International Conference on Multimedia},
  pages={7785--7794},
  year={2025}
}

@inproceedings{chen2025dove,
  title={DOVE: Efficient One-Step Diffusion Model for Real-World Video Super-Resolution},
  author={Chen, Zheng and Zou, Zichen and Zhang, Kewei and Su, Xiongfei and Yuan, Xin and Guo, Yong and Zhang, Yulun},
  booktitle = {Advances in Neural Information Processing Systems},
  pages={85218--85237},
  volume={38},
  year={2025}
}

@inproceedings{kong2025dam,
  title={Dam-vsr: Disentanglement of appearance and motion for video super-resolution},
  author={Kong, Zhe and Li, Le and Zhang, Yong and Gao, Feng and Yang, Shaoshu and Wang, Tao and Zhang, Kaihao and Kang, Zhuoliang and Wei, Xiaoming and Chen, Guanying and others},
  booktitle={Proceedings of the Special Interest Group on Computer Graphics and Interactive Techniques Conference Conference Papers},
  pages={1--11},
  year={2025}
}

@inproceedings{han2025dc,
  title={DC-VSR: Spatially and Temporally Consistent Video Super-Resolution with Video Diffusion Prior},
  author={Han, Janghyeok and Sim, Gyujin and Kim, Geonung and Lee, Hyun-Seung and Choi, Kyuha and Han, Youngseok and Cho, Sunghyun},
  booktitle={Proceedings of the Special Interest Group on Computer Graphics and Interactive Techniques Conference Conference Papers},
  pages={1--11},
  year={2025}
}

@inproceedings{sun2025one,
  title={One-step diffusion for detail-rich and temporally consistent video super-resolution},
  author={Sun, Yujing and Sun, Lingchen and Liu, Shuaizheng and Wu, Rongyuan and Zhang, Zhengqiang and Zhang, Lei},
  booktitle={Advances in Neural Information Processing Systems},
  volume={38},
  pages={172821--172841},
  year={2025}
}

@inproceedings{shi2025self,
  title={Self-supervised ControlNet with Spatio-Temporal Mamba for Real-world Video Super-resolution},
  author={Shi, Shijun and Xu, Jing and Lu, Lijing and Li, Zhihang and Hu, Kai},
  booktitle={Proceedings of the IEEE/CVF Conference on Computer Vision and Pattern Recognition},
  pages={7385--7395},
  year={2025}
}

@article{he2024venhancer,
  title={Venhancer: Generative space-time enhancement for video generation},
  author={He, Jingwen and Xue, Tianfan and Liu, Dongyang and Lin, Xinqi and Gao, Peng and Lin, Dahua and Qiao, Yu and Ouyang, Wanli and Liu, Ziwei},
  journal={arXiv preprint arXiv:2407.07667},
  year={2024}
}

@inproceedings{hong2022cogvideo,
  title={Cogvideo: Large-scale pretraining for text-to-video generation via transformers},
  author={Hong, Wenyi and Ding, Ming and Zheng, Wendi and Liu, Xinghan and Tang, Jie},
  booktitle={International Conference on Learning Representations },
  year={2023}
}

@inproceedings{yang2024cogvideox,
  title={Cogvideox: Text-to-video diffusion models with an expert transformer},
  author={Yang, Zhuoyi and Teng, Jiayan and Zheng, Wendi and Ding, Ming and Huang, Shiyu and Xu, Jiazheng and Yang, Yuanming and Hong, Wenyi and Zhang, Xiaohan and Feng, Guanyu and others},
  booktitle={International Conference on Learning Representations},
  year={2025},
}

@article{wan2025wan,
  title={Wan: Open and advanced large-scale video generative models},
  author={Wan, Team and Wang, Ang and Ai, Baole and Wen, Bin and Mao, Chaojie and Xie, Chen-Wei and Chen, Di and Yu, Feiwu and Zhao, Haiming and Yang, Jianxiao and others},
  journal={arXiv preprint arXiv:2503.20314},
  year={2025}
}

@article{zhuang2025flashvsr,
  title={Flashvsr: Towards real-time diffusion-based streaming video super-resolution},
  author={Zhuang, Junhao and Guo, Shi and Cai, Xin and Li, Xiaohui and Liu, Yihao and Yuan, Chun and Xue, Tianfan},
  journal={arXiv preprint arXiv:2510.12747},
  year={2025}
}

@inproceedings{huang2024scale,
  title={Scale-adaptive feature aggregation for efficient space-time video super-resolution},
  author={Huang, Zhewei and Huang, Ailin and Hu, Xiaotao and Hu, Chen and Xu, Jun and Zhou, Shuchang},
  booktitle={Proceedings of the IEEE/CVF Winter Conference on Applications of Computer Vision},
  pages={4216--4227},
  year={2024}
}

@inproceedings{jain2024video,
  title={Video interpolation with diffusion models},
  author={Jain, Siddhant and Watson, Daniel and Tabellion, Eric and Poole, Ben and Kontkanen, Janne and others},
  booktitle={Proceedings of the IEEE/CVF Conference on Computer Vision and Pattern Recognition},
  pages={7341--7351},
  year={2024}
}

@inproceedings{danier2024ldmvfi,
  title={Ldmvfi: Video frame interpolation with latent diffusion models},
  author={Danier, Duolikun and Zhang, Fan and Bull, David},
  booktitle={Proceedings of the AAAI Conference on Artificial Intelligence},
  volume={38},
  number={2},
  pages={1472--1480},
  year={2024}
}

@inproceedings{zhu2025generative,
  title={Generative inbetweening through frame-wise conditions-driven video generation},
  author={Zhu, Tianyi and Ren, Dongwei and Wang, Qilong and Wu, Xiaohe and Zuo, Wangmeng},
  booktitle={Proceedings of the IEEE/CVF Conference on Computer Vision and Pattern Recognition},
  pages={27968--27978},
  year={2025}
}

@inproceedings{zhang2025eden,
  title={Eden: Enhanced diffusion for high-quality large-motion video frame interpolation},
  author={Zhang, Zihao and Chen, Haoran and Zhao, Haoyu and Lu, Guansong and Fu, Yanwei and Xu, Hang and Wu, Zuxuan},
  booktitle={Proceedings of the IEEE/CVF Conference on Computer Vision and Pattern Recognition},
  pages={2105--2115},
  year={2025}
}

@inproceedings{zhang2025augmenting,
  title={Augmenting Perceptual Super-Resolution via Image Quality Predictors},
  author={Zhang, Fengjia and Rangrej, Samrudhdhi B and Aumentado-Armstrong, Tristan and Fazly, Afsaneh and Levinshtein, Alex},
  booktitle={Proceedings of the IEEE/CVF Conference on Computer Vision and Pattern Recognition},
  pages={2311--2322},
  year={2025}
}

@inproceedings{hu2021lora,
  title={LoRA: Low-Rank Adaptation of Large Language Models},
  author={Hu, Edward J and Shen, Yelong and Wallis, Phillip and Allen-Zhu, Zeyuan and Li, Yuanzhi and Wang, Shean and Wang, Lu and Chen, Weizhu},
  booktitle={International Conference on Learning Representations},
  year={2022}
}

@article{ding2020image,
  title={Image quality assessment: Unifying structure and texture similarity},
  author={Ding, Keyan and Ma, Kede and Wang, Shiqi and Simoncelli, Eero P},
  journal={IEEE Transactions on Pattern Analysis and Machine Intelligence},
  volume={44},
  number={5},
  pages={2567--2581},
  year={2020},
  publisher={IEEE}
}

@inproceedings{lai2018learning,
  title={Learning blind video temporal consistency},
  author={Lai, Wei-Sheng and Huang, Jia-Bin and Wang, Oliver and Shechtman, Eli and Yumer, Ersin and Yang, Ming-Hsuan},
  booktitle={Proceedings of the Proceedings of the European Conference on Computer Vision},
  pages={170--185},
  year={2018}
}

@inproceedings{ke2021musiq,
  title={Musiq: Multi-scale image quality transformer},
  author={Ke, Junjie and Wang, Qifei and Wang, Yilin and Milanfar, Peyman and Yang, Feng},
  booktitle={Proceedings of the IEEE/CVF International Conference on Computer Vision},
  pages={5148--5157},
  year={2021}
}

@inproceedings{wang2019edvr,
  title={Edvr: Video restoration with enhanced deformable convolutional networks},
  author={Wang, Xintao and Chan, Kelvin C.K. and Yu, Ke and Dong, Chao and Loy, Chen Change},
  booktitle={Proceedings of the IEEE/CVF Conference on Computer Vision and Pattern Recognition Workshops},
  pages={1954--1963},
  year={2019}
}

@inproceedings{zhu2019deformable,
  title={Deformable convnets v2: More deformable, better results},
  author={Zhu, Xizhou and Hu, Han and Lin, Stephen and Dai, Jifeng},
  booktitle={Proceedings of the IEEE/CVF Conference on Computer Vision and Pattern Recognition},
  pages={9300--9308},
  year={2019}
}

@inproceedings{su2017deep,
  title={Deep video deblurring for hand-held cameras},
  author={Su, Shuochen and Delbracio, Mauricio and Wang, Jue and Sapiro, Guillermo and Heidrich, Wolfgang and Wang, Oliver},
  booktitle={Proceedings of the IEEE/CVF Conference on Computer Vision and Pattern Recognition},
  pages={1279--1288},
  year={2017}
}

@inproceedings{cai2019toward,
  title={Toward real-world single image super-resolution: A new benchmark and a new model},
  author={Cai, Jianrui and Zeng, Hui and Yong, Hongwei and Cao, Zisheng and Zhang, Lei},
  booktitle={Proceedings of the IEEE/CVF International Conference on Computer Vision},
  pages={3086--3095},
  year={2019}
}

@inproceedings{tao2017detail,
  title={Detail-revealing deep video super-resolution},
  author={Tao, Xin and Gao, Hongyun and Liao, Renjie and Wang, Jue and Jia, Jiaya},
  booktitle={Proceedings of the IEEE/CVF International Conference on Computer Vision},
  pages={4472--4480},
  year={2017}
}

@inproceedings{yi2019progressive,
  title={Progressive fusion video super-resolution network via exploiting non-local spatio-temporal correlations},
  author={Yi, Peng and Wang, Zhongyuan and Jiang, Kui and Jiang, Junjun and Ma, Jiayi},
  booktitle={Proceedings of the IEEE/CVF International Conference on Computer Vision},
  pages={3106--3115},
  year={2019}
}

@inproceedings{nah2017deep,
  title={Deep multi-scale convolutional neural network for dynamic scene deblurring},
  author={Nah, Seungjun and Kim, Tae Hyun and Lee, Kyoung Mu},
  booktitle={Proceedings of the IEEE/CVF Conference on Computer Vision and Pattern Recognition},
  pages={257--265},
  year={2017}
}

@inproceedings{wang2023benchmark,
  title={Benchmark dataset and effective inter-frame alignment for real-world video super-resolution},
  author={Wang, Ruohao and Liu, Xiaohui and Zhang, Zhilu and Wu, Xiaohe and Feng, Chun-Mei and Zhang, Lei and Zuo, Wangmeng},
  booktitle={Proceedings of the IEEE/CVF Conference on Computer Vision and Pattern Recognition Workshops},
  pages={1168--1177},
  year={2023}
}

@article{loshchilov2017fixing,
  title={Fixing weight decay regularization in adam},
  author={Loshchilov, Ilya and Hutter, Frank and others},
  journal={arXiv preprint arXiv:1711.05101},
  volume={5},
  number={5},
  pages={5},
  year={2017}
}

@article{wang2004image,
  title={Image quality assessment: from error visibility to structural similarity},
  author={Wang, Zhou and Bovik, Alan C and Sheikh, Hamid R and Simoncelli, Eero P},
  journal={IEEE Transactions on Image Processing},
  volume={13},
  number={4},
  pages={600--612},
  year={2004},
  publisher={IEEE}
}

@inproceedings{zhang2018unreasonable,
  title={The unreasonable effectiveness of deep features as a perceptual metric},
  author={Zhang, Richard and Isola, Phillip and Efros, Alexei A and Shechtman, Eli and Wang, Oliver},
  booktitle={Proceedings of the IEEE/CVF Conference on Computer Vision and Pattern Recognition},
  pages={586--595},
  year={2018}
}

@article{wu2023neighbourhood,
  title={Neighbourhood representative sampling for efficient end-to-end video quality assessment},
  author={Wu, Haoning and Chen, Chaofeng and Liao, Liang and Hou, Jingwen and Sun, Wenxiu and Yan, Qiong and Gu, Jinwei and Lin, Weisi},
  journal={IEEE Transactions on Pattern Analysis and Machine Intelligence},
  volume={45},
  number={12},
  pages={15185--15202},
  year={2023},
  publisher={IEEE}
}

@inproceedings{wu2023exploring,
  title={Exploring video quality assessment on user generated contents from aesthetic and technical perspectives},
  author={Wu, Haoning and Zhang, Erli and Liao, Liang and Chen, Chaofeng and Hou, Jingwen and Wang, Annan and Sun, Wenxiu and Yan, Qiong and Lin, Weisi},
  booktitle={Proceedings of the IEEE/CVF International Conference on Computer Vision},
  pages={20087--20097},
  year={2023}
}

@inproceedings{danier2022flolpips,
  title={FloLPIPS: A bespoke video quality metric for frame interpolation},
  author={Danier, Duolikun and Zhang, Fan and Bull, David},
  booktitle={2022 Picture Coding Symposium},
  pages={283--287},
  year={2022},
  organization={IEEE}
}

@inproceedings{wang2023exploring,
  title={Exploring clip for assessing the look and feel of images},
  author={Wang, Jianyi and Chan, Kelvin CK and Loy, Chen Change},
  booktitle={Proceedings of the AAAI Conference on Artificial Intelligence},
  volume={37},
  number={2},
  pages={2555--2563},
  year={2023}
}

@inproceedings{briedis2025controllable,
  title={Controllable tracking-Based video frame interpolation},
  author={Briedis, Karlis Martins and Djelouah, Abdelaziz and Ortiz, Rapha{\"e}l and Gross, Markus and Schroers, Christopher},
  booktitle={Proceedings of the Special Interest Group on Computer Graphics and Interactive Techniques Conference Conference Papers},
  pages={1--11},
  year={2025}
}

@inproceedings{seo2025bim,
  title={BiM-VFI: Bidirectional Motion Field-Guided Frame Interpolation for Video with Non-uniform Motions},
  author={Seo, Wonyong and Oh, Jihyong and Kim, Munchurl},
  booktitle={Proceedings of the IEEE/CVF Conference on Computer Vision and Pattern Recognition},
  pages={7244--7253},
  year={2025}
}

@article{li2018video,
  title={Video super-resolution using non-simultaneous fully recurrent convolutional network},
  author={Li, Dingyi and Liu, Yu and Wang, Zengfu},
  journal={IEEE Transactions on Image Processing},
  volume={28},
  number={3},
  pages={1342--1355},
  year={2018}
}

@article{wen2022video,
  title={Video super-resolution via a spatio-temporal alignment network},
  author={Wen, Weilei and Ren, Wenqi and Shi, Yinghuan and Nie, Yunfeng and Zhang, Jingang and Cao, Xiaochun},
  journal={IEEE Transactions on Image Processing},
  volume={31},
  pages={1761--1773},
  year={2022}
}

@inproceedings{hur2025high,
  title={High-resolution frame interpolation with patch-based cascaded diffusion},
  author={Hur, Junhwa and Herrmann, Charles and Saxena, Saurabh and Kontkanen, Janne and Lai, Wei-Sheng and Shih, Yichang and Rubinstein, Michael and Fleet, David J and Sun, Deqing},
  booktitle={Proceedings of the AAAI Conference on Artificial Intelligence},
  volume={39},
  number={4},
  pages={3868--3876},
  year={2025}
}

@inproceedings{xu2025videogigagan,
  title={Videogigagan: Towards detail-rich video super-resolution},
  author={Xu, Yiran and Park, Taesung and Zhang, Richard and Zhou, Yang and Shechtman, Eli and Liu, Feng and Huang, Jia-Bin and Liu, Difan},
  booktitle={Proceedings of the IEEE/CVF Conference on Computer Vision and Pattern Recognition},
  pages={2139--2149},
  year={2025}
}

@inproceedings{wang2018occlusion,
  title={Occlusion aware unsupervised learning of optical flow},
  author={Wang, Yang and Yang, Yi and Yang, Zhenheng and Zhao, Liang and Wang, Peng and Xu, Wei},
  booktitle={Proceedings of the IEEE/CVF Conference on Computer Vision and Pattern Recognition},
  pages={4884--4893},
  year={2018}
}

@inproceedings{zhang2025continuous,
  title={Continuous Space-Time Video Resampling with Invertible Motion Steganography},
  author={Zhang, Yuantong and Chen, Zhenzhong},
  booktitle={Proceedings of the IEEE/CVF Conference on Computer Vision and Pattern Recognition},
  pages={2116--2126},
  year={2025}
}

@article{zhang2022optical,
  title={Optical flow reusing for high-efficiency space-time video super resolution},
  author={Zhang, Yuantong and Wang, Huairui and Zhu, Han and Chen, Zhenzhong},
  journal={IEEE Transactions on Circuits and Systems for Video Technology},
  volume={33},
  number={5},
  pages={2116--2128},
  year={2023}
}

@inproceedings{becker2025continuous,
  title={Continuous Space-Time Video Super-Resolution with 3D Fourier Fields},
  author={Becker, Alexander and Erbach, Julius and Narnhofer, Dominik and Schindler, Konrad},
  booktitle={International Conference on Learning Representations},
  year={2026}
}

@inproceedings{yin2024one,
  title={One-step diffusion with distribution matching distillation},
  author={Yin, Tianwei and Gharbi, Micha{\"e}l and Zhang, Richard and Shechtman, Eli and Durand, Fredo and Freeman, William T and Park, Taesung},
  booktitle={Proceedings of the IEEE/CVF Conference on Computer Vision and Pattern Recognition},
  pages={6613--6623},
  year={2024}
}

@inproceedings{wang2024sinsr,
  title={Sinsr: diffusion-based image super-resolution in a single step},
  author={Wang, Yufei and Yang, Wenhan and Chen, Xinyuan and Wang, Yaohui and Guo, Lanqing and Chau, Lap-Pui and Liu, Ziwei and Qiao, Yu and Kot, Alex C and Wen, Bihan},
  booktitle={Proceedings of the IEEE/CVF Conference on Computer Vision and Pattern Recognition},
  pages={25796--25805},
  year={2024}
}

@inproceedings{lin2025diffusion,
  title={Diffusion Adversarial Post-Training for One-Step Video Generation},
  author={Lin, Shanchuan and Xia, Xin and Ren, Yuxi and Yang, Ceyuan and Xiao, Xuefeng and Jiang, Lu},
  booktitle={Proceedings of the 42nd International Conference on Machine Learning},
  pages={37959--37974},
  year={2025},
  organization={PMLR}
}

@inproceedings{liu2023instaflow,
  title={Instaflow: One step is enough for high-quality diffusion-based text-to-image generation},
  author={Liu, Xingchao and Zhang, Xiwen and Ma, Jianzhu and Peng, Jian and others},
  booktitle={International Conference on Learning Representations},
  year={2024}
}

@inproceedings{sun2025pixel,
  title={Pixel-level and semantic-level adjustable super-resolution: A dual-lora approach},
  author={Sun, Lingchen and Wu, Rongyuan and Ma, Zhiyuan and Liu, Shuaizheng and Yi, Qiaosi and Zhang, Lei},
  booktitle={Proceedings of the IEEE/CVF Conference on Computer Vision and Pattern Recognition},
  pages={2333--2343},
  year={2025}
}

@inproceedings{marealtime,
  title={Realtime Video Frame Interpolation using One-Step Diffusion Sampling},
  author={Ma, Yongrui and Zhao, Shijie and Yao, Mingde and Li, Junlin and Liu, Xiaohong and Dou, Qi and Gu, Jinwei and Xue, Tianfan and others},
  booktitle={International Conference on Learning Representations},
  year={2026}
}

@article{ho2022video,
  title={Video diffusion models},
  author={Ho, Jonathan and Salimans, Tim and Gritsenko, Alexey and Chan, William and Norouzi, Mohammad and Fleet, David J},
  journal={Advances in Neural Information Processing Systems},
  volume={35},
  pages={8633--8646},
  year={2022}
}

@article{zhang2023i2vgen,
  title={I2vgen-xl: High-quality image-to-video synthesis via cascaded diffusion models},
  author={Zhang, Shiwei and Wang, Jiayu and Zhang, Yingya and Zhao, Kang and Yuan, Hangjie and Qin, Zhiwu and Wang, Xiang and Zhao, Deli and Zhou, Jingren},
  journal={arXiv preprint arXiv:2311.04145},
  year={2023}
}
